\pdfoutput=1

\documentclass[11pt]{article}

\usepackage[final]{acl}

\usepackage{times}
\usepackage{latexsym}

\usepackage[T1]{fontenc}

\usepackage[utf8]{inputenc}

\usepackage{microtype}

\usepackage{graphicx}

\usepackage{hyperref}
\usepackage{url}
\usepackage{booktabs}
\usepackage{amsfonts}
\usepackage{nicefrac}
\usepackage{microtype}
\usepackage{xcolor}
\usepackage{graphicx}
\usepackage{enumitem}
\usepackage{bm}
\usepackage{multirow}
\usepackage{tabularx}
\usepackage{adjustbox}
\usepackage{algorithm}
\usepackage{algpseudocode}
\usepackage{ifthen}
\usepackage{subcaption}
\usepackage{comment}
\usepackage{amsmath}
\usepackage{natbib}
\usepackage{listings}

\lstdefinestyle{mystyle}{
  basicstyle=\ttfamily\small,
  breaklines=true,
  frame=single,
  backgroundcolor=\color{gray!5},
  columns=fullflexible,
  keepspaces=true
}
\definecolor{darkblue}{rgb}{0, 0, 0.5}
\hypersetup{colorlinks=true, citecolor=darkblue, linkcolor=darkblue, urlcolor=darkblue}

\newboolean{showcomments}
\setboolean{showcomments}{false}  

\newboolean{hideaddressedcomments}
\setboolean{hideaddressedcomments}{false}  

\newcommand{\showcomment}[3]{%
    \ifthenelse{\boolean{showcomments}}
        {\colorbox{#1}{\textcolor{white}{#2}}~{\textcolor{#1}{#3}}}
        {} 
}

\newcommand{\addressed}[1]{%
    \ifbool{hideaddressedcomments}{}{#1}%
}

\newcommand{\jiateng}[1]{\showcomment{MyDarkBlue}{jiateng}{#1}}

\newcommand{\yuji}[1]{\showcomment{orange}{yuji}{#1}}

\newcommand{\yi}[1]{\showcomment{teal}{yi}{#1}}

\newcommand{\uky}[1]{\showcomment{byzantine}{uky}{#1}}

\newcommand{\ignore}[1]{}

\makeatletter
\DeclareRobustCommand\onedot{\futurelet\@let@token\@onedot}
\def\@onedot{\ifx\@let@token.\else.\null\fi\xspace}

\makeatother

\definecolor{MyDarkBlue}{rgb}{0,0.08,0.8}
\definecolor{MyDarkGreen}{RGB}{45,155,45}
\definecolor{MyDarkRed}{rgb}{0.8,0.02,0.02}
\definecolor{MyOrange}{rgb}{1.0, 0.4, 0.2}
\definecolor{MyPurple}{RGB}{111,0,255}
\definecolor{MyRed}{rgb}{0.8,0.0,0.0}
\definecolor{MyGold}{rgb}{0.75,0.6,0.12}
\definecolor{MyDarkgray}{rgb}{0.66, 0.66, 0.66}

\definecolor{RoseQuartzBg}{HTML}{F7CAC9}
\definecolor{RoseQuartz}{HTML}{F5A798}
\definecolor{Serenity}{HTML}{92A8D1}
\definecolor{OrangeRed}{rgb}{1.0, 0.27, 0.0}
\definecolor{Turquoise}{HTML}{0F4C81}
\definecolor{mint}{rgb}{0.24, 0.71, 0.54}
\definecolor{byzantine}{rgb}{0.74, 0.2, 0.64}
\definecolor{byzantium}{rgb}{0.44, 0.16, 0.39}
\definecolor{captioningtask}{HTML}{9C843F}
\definecolor{qatask}{HTML}{CC6600}
\definecolor{temporalmarker}{HTML}{7F00FF}
\definecolor{targettext}{HTML}{3333FF}
\definecolor{prompttext}{HTML}{666666}
\definecolor{videolevel}{HTML}{330066}
\definecolor{framelevel}{HTML}{0066CC}
\definecolor{tokenlevel}{HTML}{336600}
\definecolor{boxgrey}{HTML}{666666}
\definecolor{boxblue}{HTML}{6C8EBF}
\definecolor{boxgreen}{HTML}{82B366}
\definecolor{textgreen}{HTML}{009900}
\definecolor{textred}{HTML}{FF0000}
\definecolor{textreddark}{HTML}{CC0000}
\definecolor{textblue}{HTML}{0066CC}

\title{EMCompress: Video-LLMs with Endomorphic Multimodal Compression}

\author{Zheyu Fan\thanks{Work done during internship at UIUC.}$^{1,2}$, Jiateng Liu$^{1}$, Yuji Zhang$^{1}$, Zihan Wang$^{2}$, \\
{\bfseries Yi R. (May) Fung$^{1}$, Manling Li$^{2}$, Heng Ji$^{1}$} \\[0.3em]
$^{1}$University of Illinois Urbana-Champaign  ~~~~$^{2}$Northwestern University\\
\texttt{zheyufan@u.northwestern.edu} ~~~ \texttt{hengji@illinois.edu}
}

\begin{document}

\maketitle

\begin{abstract}

Video-LLMs face a fundamental tension in long-video reasoning: static, sparse frame sampling either dilutes evidence across task-irrelevant segments at significant cost or misses fine-grained temporal semantics altogether. We propose a novel, cognitively-inspired task --- \underline{\textbf{E}}ndomorphic \underline{\textbf{M}}ultimodal \underline{\textbf{C}}ompression (EMC) --- as a \textit{structurally-constrained sufficient-statistic problem} for VideoQA, and formulate it as an endomorphic transformation $\mathcal{F}_{EMC}:(V,Q) \to (v,q)$ that compresses the multimodal input while preserving answer invariance across reasonable downstream models. The endomorphic form keeps the compressed output in the downstream pipeline's native task space --- a structural mirror of the filter-then-reason mechanism in the cognitive literature motivating EMC --- distinguishing it from latent-code compression (IB / VIB) and making the formulation extensible to other multimodal settings. Under the Markov chain $A \to (V,Q) \to (v,q)$, EMC realizes the classical sufficiency condition $I((v,q); A) = I((V,Q); A)$ in its VideoQA-natural form. As a modular front-end, EMC plugs into both Video Instruction Tuning and Video Question Answering pipelines. We release the first dedicated benchmark and propose ReSimplifyIt, an EMC baseline surpassing prior methods by 0.40 F-1 with competitive query rewriting. Integrating EMC yields relative gains of 7.33\% in training and 33.7\% in inference for video-language understanding.\footnote{Our code is available \href{https://github.com/LordUky/Temporal-Visual-Screening}{here}.}

\end{abstract}

\section{Introduction}

\begin{figure*}[t]
  \includegraphics[width=\textwidth]{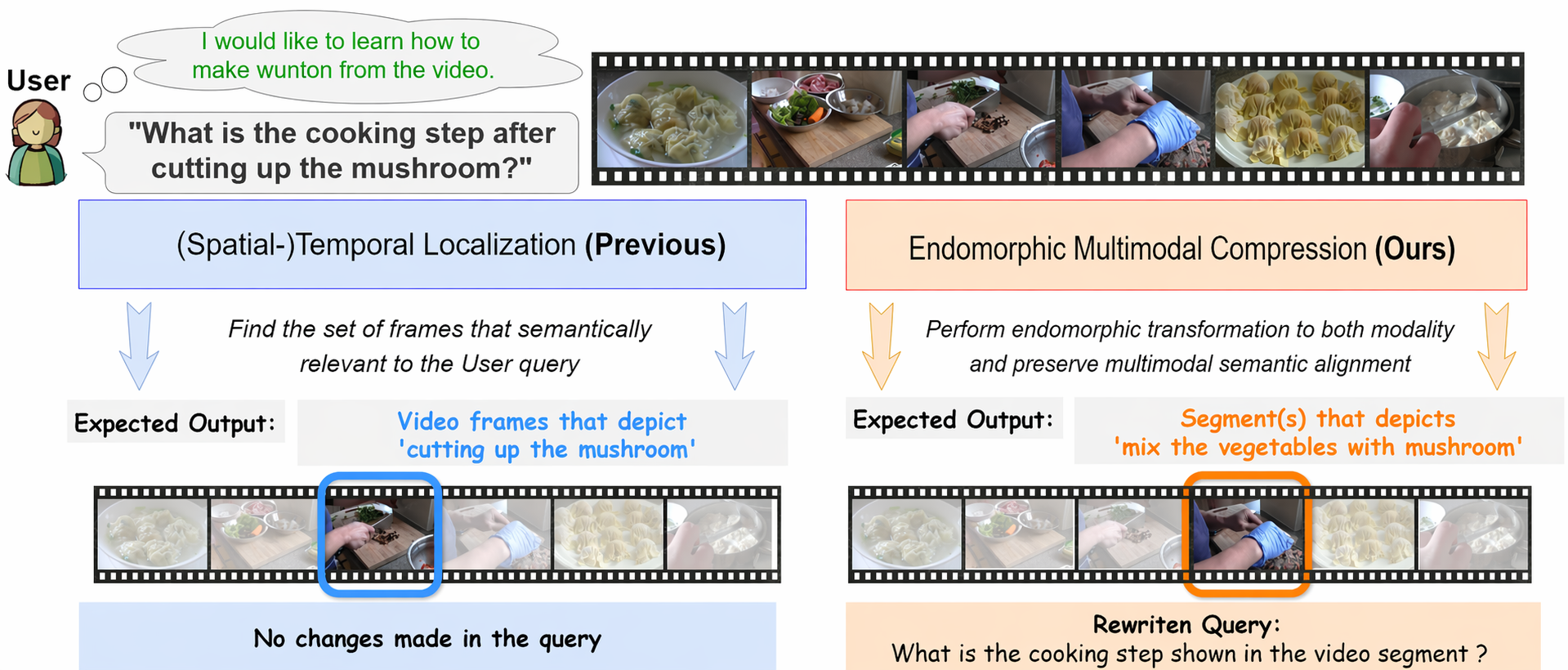}
  \caption{Input and output example of the EMC task as a side-by-side comparison to a superficially similar task: temporal localization. EMC focuses on goal-driven cognitive alignment and performs reasoning-guided problem reconstruction through information compression, whereas grounding only emphasizes perceptual alignment and locates depictive level visual evidence by direct moment retrieval.\vspace{-10pt}}
  \label{fig:emc_task_compare}
\end{figure*}

Human cognition optimizes information processing via a two-stage control mechanism: a fast, pre-attentive filtering to prune redundancy, followed by focused reasoning on a compacted, goal-aligned representation \citep{FITNO1, GS2.0NO2, zacks2007event, topdownfeedback}. Behaviorally, this manifests as scrubbing the seek bar before detailed viewing \citep{wu2024v}, minimizing Extraneous Load to purify Germane Load \citep{cognitiveLoadDuringProblemSolving:EffectsOnLearningNO3, NineWaysToReduceCognitiveLoadInMultimediaLearningNO4, ReducingExtraneousLoadImportantNO15}---reshaping the problem before engaging the expensive reasoning stage.

This filter-before-reason economy is forced by VideoQA: a minute of video contains thousands of frames of which only a narrow band bears on any given query, yet current Video-LLMs \citep{chen2023video-LLMmodelingvideosequence,2023videochat,Xu2024PLLaVAP,maaz-etal-2024-video,10658165,Ma_2024_CVPR} typically fit the entire stream through a bounded visual-token budget via query-agnostic uniform sampling. This dilutes semantic signal at critical segments and injects noise, weakening supervision during training \citep{lin2024norton, knowledgeovershadowing} and reducing task-oriented focus at inference.

Existing work such as traditional grounding and grounded QA \citep{nextgqa, cgbench} attempts the obvious remedy of retrieving query-relevant frames, but operates at the \textbf{perceptual level}---depictive similarity between query tokens and visual evidence. Complex queries entangle multi-hop temporal, relational, and causal dependencies with sparse, weakly localized cues, so retrieving depicted frames without adapting the query leaves a multimodal semantic mismatch that misleads the downstream model.

This calls for a novel mechanism that jointly reshapes reasoning structure and cognitive load while treating input modalities \textbf{as a whole}. The two-stage structure of human cognition \citep{FITNO1, GS2.0NO2} suggests a clean architectural response: decouple filtering from reasoning so one mechanism serves arbitrary downstream reasoners. We design a \textbf{front-end adapter} that reshapes the raw instance symmetrically over all modalities before it reaches any Video-LLM.

The cited cognitive mechanisms share a structural property: the filter stage's output stays within the substrate the reasoning stage natively consumes---Feature Integration masks the visual field in place \citep{FITNO1}, Guided Search overlays priority on the original perceptual map \citep{GS2.0NO2}, event segmentation carves segments out of the ongoing temporal stream \citep{zacks2007event}; none transcode to an alien representation. Taking this commitment seriously forces the computational filter into the downstream reasoner's native task space---that is, to be \textbf{endomorphic}:
\[
\mathcal{F}_{\text{EMC}} : (V, Q) \rightarrow (v, q),
\]
mapping a multimodal task instance back into the same task space. This distinguishes EMC from latent-code compression (Information Bottleneck / VIB), whose learned code is not natively consumable by existing pipelines. Model-agnostic answer invariance---formalized as C2 in \S\ref{emc_def}---is a separate behavioral requirement parallel to endomorphism, not derived from it. The endomorphic form additionally makes the formulation extensible to \textbf{any multimodal task} admitting a uniform agent signature; VideoQA is one instance.

Taking a generative view---where answer $A$ is a latent fact and $(V, Q)$ is an observation drawn from it---this compression forms the Markov chain $A \to (V, Q) \to (v, q)$, and \underline{\textbf{E}}ndomorphic \underline{\textbf{M}}ultimodal \underline{\textbf{C}}ompression (EMC) is precisely a \textbf{structurally-constrained sufficient-statistic problem}: find the most compact admissible $(v, q)$ preserving all $A$-relevant information. This identification places EMC in the Information Bottleneck lineage \citep{tishby1999information, tishby2015deep}, instantiated over the original multimodal input space rather than a learned latent code. $A$ is conserved throughout, aligning with the fixity of the Goal State in Problem Space Theory \citep{HumanProblemSolvingNO12}: EMC reshapes ``what to ask'', not ``what to answer''.

EMC yields four concrete benefits: (i) Training alignment: removing off-target supervision boosts gradient signal-to-noise and reduces spurious correlations under fixed budgets; (ii) Inference robustness: offloading non-informative perceptual burdens shortens the implicit reasoning program, unleashing reasoning capabilities; (iii) Interpretability: the compact $(v, q)$ is a controllable, faithful artifact exposing model bottlenecks; (iv) Generalizability: as a model-agnostic front-end, EMC benefits diverse VideoQA agents.

We present \textit{ReSimplifyIt}, a plug-and-play multi-agent framework that \underline{\textbf{ReS}}iliently \underline{\textbf{imp}}rovises and qua\underline{\textbf{lif}}ies proposals \underline{\textbf{It}}eratively. Each compression round runs a language-only Launcher that hypothesizes a trimming instruction and rewritten query (using the intrinsic cross-modal relation as a prior), a Validator that executes and critiques these plans through a Viewer module, and memory trackers for self-correction---operationalizing a competence-from-consequence~\citep{brooks1991intelligence} principle: trial execution provides constructive feedback tightening the coupling between hypothesized reasoning structure and available evidence.

We further construct EMCompress, the first benchmark dedicated to evaluating EMC as a joint transformation over video and query. Benchmarking shows EMC is a non-trivial task far from solved, revealing a key bottleneck for downstream VideoQA reasoning and deserving independent study.

To our knowledge, ours is the first work to formally recognize that grounding and trimming a video creates a semantic mismatch with complex, context-dependent queries, to formalize the task on this premise, and to extensively explore temporal filtering within the Video-LLM framework. Our results show that EMC not only improves inference-time performance but also enables more structured and effective training through finer-grained multimodal alignment, paving the way for scalable, interpretable, cognitively-inspired video-language understanding systems.

The main contributions of this work are summarized below:

\begin{itemize}[leftmargin=*,topsep=-1pt,itemsep=0ex]
\item We propose endomorphic multimodal compression (EMC), a novel cognitively-inspired, information-theoretically framed task that guides task-aware reasoning structure filtering and addresses a key bottleneck in multi-modal alignment. We also release EMCompress, a benchmark with 238.2 hours of video and 2{,}754 questions.
\item We introduce ReSimplifyIt, the first baseline for the EMC task: a model-agnostic plug-in compatible with \textbf{any} existing videoQA or instruction-tuning pipeline.
\item We quantitatively evaluate EMC in VideoQA and Video Instruction Tuning, where our method delivers over $10\%$ and $5\%$ absolute gains at inference and training, respectively, across multiple strong baselines and benchmarks.
\end{itemize}

\begin{figure*}[t]
  \includegraphics[width=\textwidth]{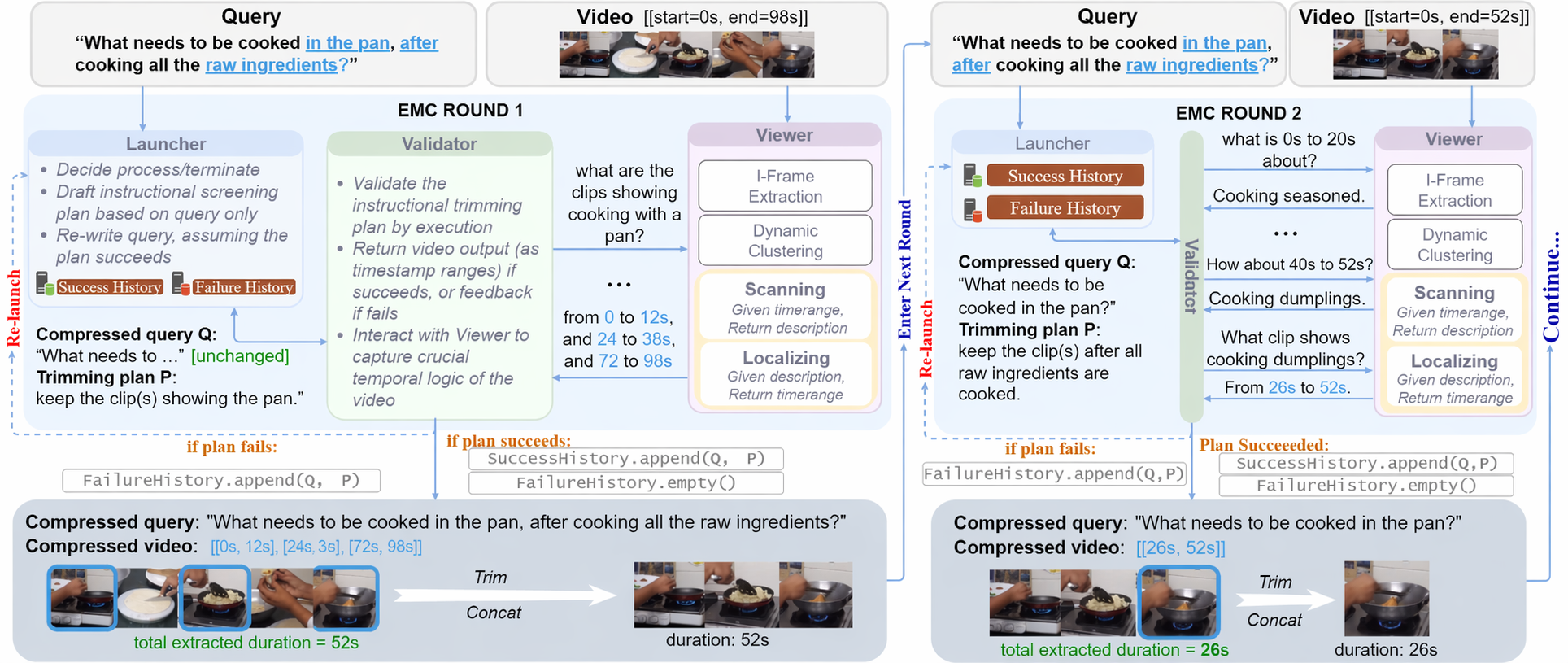}  
  \caption{Snapshot examples of the workflow of our proposed ReSimplifyIt framework.}
  \label{fig:workflow_figure}
\end{figure*}

\section{Formulation of Endomorphic Multimodal Compression (EMC)}

\subsection{Task Definition}
\label{emc_def}
We formulate Endomorphic Multimodal Compression (EMC) as a \textbf{structurally-constrained compression problem} grounded in a generative view of VideoQA. We treat the ground-truth answer $A$ as a latent fact about the world---e.g., a procedural truth such as \textit{``the step after cutting mushrooms is mixing vegetables''}---with the video $V$ and query $Q$ jointly observed as $(V, Q) \sim p(V, Q \mid A)$. A video records an instantiation of the underlying fact; a natural-language query points to an aspect thereof. This mirrors the measurement-invariance principle in the physical sciences: observations \textit{access} a fact rather than constitute it.

Under this view, any compression $(V, Q) \to (v, q)$ forms the Markov chain
\vspace{-0.2em}
\[
A \;\to\; (V, Q) \;\to\; (v, q),
\]
and the Data Processing Inequality~\citep{cover2006elements} bounds the mutual information $I(\cdot\,;\cdot)$ as
\vspace{-0.2em}
\[
I\big((v, q);\, A\big) \;\leq\; I\big((V, Q);\, A\big),
\]
with equality iff $(v, q)$ is an $A$-sufficient statistic of $(V, Q)$.\footnote{See Appendix~\ref{appendix:measurability} for discussion.} EMC is the problem of finding such a sufficient statistic under additional structural and VideoQA-specific constraints: temporal continuity on $v$, answer invariance at instance level across reasonable downstream models, and joint minimality on both modalities. This situates EMC within the \textit{Information Bottleneck lineage}~\citep{tishby1999information, tishby2015deep, alemi2017deep}, instantiated over the original multimodal input space.

Formally, EMC defines a transformation
\vspace{-0.2em}
\[
\mathcal{F}_{\text{EMC}}:\; (V, Q) \;\longmapsto\; (v, q)
\]
over the original multimodal task space (Figure~\ref{fig:emc_task_compare}), where $(v, q)$ is required to satisfy two \textit{admissibility conditions} and minimize two complementary \textit{complexity measures}.

\paragraph{Admissibility conditions.}
\begin{itemize}[leftmargin=*,topsep=2pt,itemsep=1pt]
    \item \textbf{(C1) Structural Continuity.} $v$ is the concatenation of $n \geq 1$ non-overlapping contiguous subsegments of $V$:
    \vspace{-0.2em}
    \[
    v \;=\; \bigcup\nolimits_{i=1}^{n} \{F_{s_i}, \ldots, F_{e_i}\} \;\subseteq\; V,
    \]
    with $1 \leq s_1 \leq e_n \leq T$, preserving low-level temporal integrity for motion, appearance, and scene dynamics.

    \item \textbf{(C2) Answer Sufficiency.} For any reasonable VideoQA agent $\mathcal{M}$,
    \vspace{-0.2em}
    \[
    \mathcal{M}(q, v) \;=\; \mathcal{M}(Q, V).
    \]
    This is the VideoQA-natural form of the sufficient-statistic identity $I((v,q); A) = I((V,Q); A)$: VideoQA has no distribution-averaged notion of correctness---every instance has a specific ground-truth answer that the user expects the model to return, so sufficiency must hold at instance level across reasonable downstream models rather than merely on average. Under this instantiation, (C2) implies the MI identity by preserving the full distribution of model outputs,\footnote{See Appendix~\ref{appendix:mi_answer_link} for discussion.} locating EMC in the sufficient-statistic regime while specifying its VideoQA-specific form.
\end{itemize}

\noindent Let $\mathcal{A}$ denote the set of pairs $(v, q)$ satisfying both (C1) and (C2).

\paragraph{Minimality objectives.}
\begin{itemize}[leftmargin=*,topsep=2pt,itemsep=1pt]
    \item \textbf{(O1) Video-side minimality.} Minimize $\mathrm{Size}(v)$, the content retained from $V$ (e.g., total duration or frame count).
    \item \textbf{(O2) Query-side minimality.} Minimize $\mathrm{Infer}(q)$, the number of reasoning steps required to derive $A$ from $(v, q)$. For example, reasoning on \textit{``within 5\,s after event A''} incurs three steps: localizing event A, relational reasoning on \textit{``after''}, and temporal reasoning on \textit{``5\,s''}.
\end{itemize}

Rather than tracing a Pareto frontier, the two objectives admit a unique solution via \textbf{video-priority lexicographic} resolution\footnote{A mechanism-level account---based on \textit{constraint re-allocation} between modalities---of why the two objectives collapse to a single operating point rather than a Pareto continuum is provided in Appendix~\ref{appendix:pareto_collapse}.}: video compression drives the transformation and query adaptation is its downstream consequence. Formally,
\vspace{-0.3em}
\[
(v^*, q^*) = \Big(\!\!\operatorname*{arg\,min}_{\substack{v\,:\,\exists\, q,\\ (v,q)\in\mathcal{A}}}\!\mathrm{Size}(v),\ \operatorname*{arg\,min}_{q\,:\,(v^*, q)\in\mathcal{A}}\!\mathrm{Infer}(q)\Big).
\]

\subsection{Integrating EMC into Video-LLM Workflows}
\label{subsec:emc_integration}

Because $\mathcal{F}_{\text{EMC}}$ is endomorphic, any existing Video-LLM training or inference pipeline continues to operate unchanged, receiving $(v, q)$ in place of $(V, Q)$; see Appendix~\ref{appendix:emc_integration} for details.




\jiateng{I feel the best way to present this is that we first have a section enhancing that 'EMC is the bottleneck of current Video-LLMs', say we assume that EMC tasks can be perfectly resolved, we show a table where model performance can be significantly boosted. However, we also show that current appraoches does not resolve EMC well. We use this section to enhance the importance of the EMC task itself, before coming to an approach that 'solves EMC' } \uky{added leading words for eval results section, to proper refer to the tables}

\section{Method}
\label{sec:method}

\addressed{
\jiateng{General Comments: We need to make sure that the design of each module is of novelty and insightful. The current writing did not highlight these. We need to highlight why the design is good and how is it different from exisitng framework.} \uky{done}

\jiateng{For Launcher: It is different from common planners due to its availability to check history and can be improved with feedback for multiple times} \uky{done}

\jiateng{For Validator: Helps verify the planning, making sure everything is going on smoothly, also serves as an interface to the viewer} \uky{done}

\jiateng{For Viewer: Introduce why the two core ability is the most important, and how the implementation is helpful.} \uky{done}
}

\addressed{\jiateng{At the same time, the current intro to the framework is way too long. Now it takes up tp 4 pages, we need to shorted this to at most 2 pages (including your algorithms.) We need writings to emphazies on the training, and derive some qualitative analysis for both stages.}}

\subsection{ReSimplifyIt Framework}\yuji{section 3.1 is much clearer than section 2. Could you write section 2.1 in this style?}
Drawing inspiration from \textit{dynamic and interactive attention deployment} \citep{FITNO1, GS2.0NO2} and humans' information compression process by dragging the video's progress bar, we propose our multi-agent ReSimplifyIt framework. The framework is composed of three main agentic components: the Launcher, the Validator, and the Viewer, accompanied by two extra helper modules as memory trackers: the Failure History and the Success History. For implementation details, please refer to Section \ref{sec:experiments}. We denote $\{(v_r, q_r)|1 \leq r \leq R\}$ as the whole EMC process, where $v_r, q_r$ represent the output video clip and question of the $r$th round, respectively, and $R$ sets the maximum number of rounds. Refer to Algorithm \ref{alg_emc} for the complete algorithm of our EMC process performed by our ReSimplifyIt framework, and Figure \ref{fig:workflow_figure} for a snapshot demonstration. 

\paragraph{Initiating an EMC Round: Launcher drafting a video trimming instruction}

The EMC process unfolds over iterative rounds, akin to how humans drag the progress bar multiple times based solely on empirical surmise before actually viewing the video content. In each round~$r$, a \textbf{Launcher} module generates a trimming instruction~$i_r$ and a revised query~$q_r$ as a trial, where the trimming instruction may only contain \textit{high-level} semantics yielding a declarative goal (e.g. ``keep the clip after event X'' instead of ``clip([5s,10s])''), based solely on the previous query~$q_{r-1}$, without vision access. This trial-based methodology strikes an effective balance between performance and efficiency as the inherent semantic relevance between the query and video input can thereby serve as a prior to enabling plausible instruction generation without video access.

Akin to a human landing on undesired segments after dragging the progress bar, we equip the Launcher module with self-correction based adaptive refinement mechanisms to tackle rare failure cases and ensure correctness. Specifically, we maintain two memory tracker modules: \textbf{Success History (SH)} and \textbf{Failure History (FH)}. A successful trial is added to SH and terminates the round; otherwise, it is recorded in FH and retriggers the Launcher with updated feedback.

We formalize the Launcher's behavior as:
\begin{center}
$(i_r, q_r) = \texttt{Launcher}(q_{r-1}, FH, SH, t)$
\end{center}
where $t$ denotes the task prompt.

Our feedback-driven formulation achieves iterative self-correction, a key distinction from prior work~\citep{VideoAgent, shang-etal-2024-traveler}. It decomposes the task into progressive rounds to enable iterative self-correction and foster structured reasoning, which enhances multi-hop robustness while minimizing correction costs. The lightweight, video-free Launcher further boosts efficiency by performing abductive reasoning on the language query without sacrificing fidelity.

\paragraph{Validating a trial: Validator executing instruction}

The \textbf{Validator module} receives the high-level instruction $i_r$ from the Launcher and determines its success through execution. The module performs two tightly coupled tasks:  
(i) \textit{assessing the feasibility} of the instruction (i.e., whether it is able to succeed), and  
(ii) \textit{executing} the instruction to obtain results or feedback.
Unlike the Launcher, the Validator \textit{has indirect access to visual semantics} by interacting with the Viewer module while not taking visual inputs or frame captions itself.

It returns a tuple $(d_r, m_r)$, where:
\begin{itemize}
  \item $d_r$ indicates whether the instruction is deemed \texttt{succeeded} (feasible) or \texttt{failed} (infeasible),
  \item $m_r$ contains either the resulting trimmed video (in the form of timestamp ranges) if successfully executed the instruction, or an explanation if failed.
\end{itemize}

\begin{center}
$(d_r, m_r) = \texttt{Validator}(i_r, v_{r-1}, t)$
\end{center}

Here, $\texttt{Validator}()$ denotes the Validator module, $v_{r-1}$ is the previous video state, and $t$ is the task prompt. 

The Validator may also decide to call the Viewer module before returning to Launcher. In this scenario, $d_r$ and $m_r$ stand for this decision symbol and the message to the Viewer, respectively.

Our unified Validator assesses plan viability via direct execution rather than handcrafted rules, embodying a ``competence from consequence'' philosophy~\citep{brooks1991intelligence}. This integration simplifies control flow and reduces inter-module communication. Crucially, failures are not terminal but provide constructive feedback for the Launcher to refine instructions, transforming execution into a mechanism for both validation and continual adaptation.

\paragraph{Scanning the video: Viewer scanning and localizing}

Inspired by the Bottom-Up (stimulus-driven) and Top-Down (goal-driven) Activation schema proposed by GS2.0 \citep{GS2.0NO2}, which also guides humans' scrubbing the seek bar when watching videos, we design the \textbf{Viewer} module to explicitly incorporate two complementary tasks: 

\textit{\textbf{(i) Scanning:}} summarize the content of a video snippet given a timestamp range;

\textit{\textbf{(ii) Localizing:}} retrieve a timestamp range given a text summarization of a video snippet.

With this complementarily symmetric design, the Viewer module achieves elegant and efficient bidirectional navigation by providing both top-down (content-driven) and bottom-up (time-driven) exploration, making the Viewer module highly flexible.

To support both tasks, we first perform a two-stage keyframe extraction and captioning process: (1) extract I-frames using MPEG-4 compression to capture key visual content, and (2) apply a dynamic frame clustering algorithm to select the final keyframes, which adaptively adjusts the number of clusters without supervision. Compared to frame clustering techniques in previous work \citep{wang2024videotreeadaptivetreebasedvideo}, this approach demonstrates stronger generalizability and robustness. For the algorithm of the keyframe extraction process, please refer to algorithm \ref{alg_isodata}. This pre-processing is denoted as:

\begin{center}
$C = prep(start, end).$
\end{center}

\textbf{Scanning} executes by summarizing the snippet content via LLM reasoning over $C$, optionally querying additional frames, and is denoted as:
\begin{center}
$Cap = Scanner(prep(start, end), t).$
\end{center}

\textbf{Localizing} follows a lightweight three-stage search: (1) locate top-k candidate timestamps, (2) select the best, and (3) expand it into a full range. The LLM may optionally query extra frames for confirmation, ensuring minimal frame access while maintaining accuracy. This process is denoted as:
\begin{center}
$(t_{start}, t_{end}) = Localizer(prep(0, d), q).$
\end{center}

This lightweight yet effective three-stage design achieves fine-grained temporal grounding with minimal overhead, showcasing the Viewer’s adaptability and plug-and-play potential.

Algorithms of the keyframe extraction and localization stages are provided in appendix \ref{appendix_viewer}.

\subsection{EMC-guided inference}
\label{subsection:emc-guided-inference}

Attributed to its endomorphic property, EMC can serve as a plug-and-play adapter process for any VideoQA pipeline, including both inference and training stages. Given the input question $Q$ and video $V$, EMC-guided inference is conducted as:

$$ response = VideoQA(v, q) $$
$$ v, q = \mathcal{F}_{\text{EMC}}(V, Q) $$

where $VideoQA$ can be \emph{any} VideoQA pipeline, including video-LLMs, LLM-assisted pipelines, or any others alternatives, and $\mathcal{F}_{\text{EMC}}$ stands for the Endomorphic Multimodal Compression process.

\subsection{EMC-guided training}
\label{subsection:emc-guided-training}
EMC can also be applied to training stage to purify supervision signal. Given input question $Q$, input video $V$, and ground truth answer $a$, we compute the likelihood during the EMC-guided training as:

\vspace{-2em}




\begin{small}
$$
p\left( \mathbf{X}_\mathrm{A} \mid \mathbf{X}_\mathrm{V}, \mathbf{X}_\mathrm{Q} \right)
= \prod_{i=1}^{L} p_{\theta} \left( \mathbf{X}_\mathrm{A}^{[i]} \mid \mathbf{X}_\mathrm{V}, \mathbf{X}_\mathrm{Q}, \mathbf{X}_\mathrm{A}^{[1:i-1]} \right)
$$ 

$$ \mathbf{X}_\mathrm{A}, \mathbf{X}_\mathrm{Q}, \mathbf{X}_\mathrm{V} = f_t(a), f_t(q), f_v(v)$$

$$ v, q = \mathcal{F}_{\text{EMC}}(V, Q) $$
\end{small}

where $f_t$, $f_v$ are the text and visual tokenizers. \addressed{\yi{this last phrase is repeated across the last two subsections. Math formulation can also be better explained imo.}\uky{unnecessary phrasing removed, and math formula repolished}}

\section{Experiment Settings}
\label{sec:experiments}

\begin{table*}[t]
  \centering

    \begin{subtable}{\textwidth}
      \centering
      \resizebox{\textwidth}{!}{
      \small
        \begin{tabular}{c|cc|cc|cc|cc}
          \toprule
          \textbf{\multirow{2}{*}{Method}}  & \multicolumn{2}{c}{Temporal Relational} & \multicolumn{2}{c}{Timepoint Indexed} & \multicolumn{2}{c}{Multifaceted Integrative} & \multicolumn{2}{c}{\textbf{Average}} \\
           & mIoU & F1 & mIoU & F1 & mIoU & F1 & mIoU & F1 \\
          \midrule
          VTG-GPT \citep{vtggpt} & \underline{0.17} & \underline{0.29} & \underline{0.15} & \underline{0.26} & \underline{0.16} & \underline{0.24} & \underline{0.16} & \underline{0.27} \\
          \citet{tfvtg} & 0.11 & 0.19 & 0.04 & 0.08 & 0.06 & 0.10 & 0.07 & 0.12 \\
          \textbf{ReSimplifyIt} (Ours) & \textbf{0.23} & \textbf{0.37} & \textbf{0.98} & \textbf{0.99} & \textbf{0.47} & \textbf{0.64} & \textbf{0.56} & \textbf{0.67} \\
          \bottomrule
        \end{tabular}
      }
      \caption{Results on video output}
      \label{tab:emc_main_reduced_vid}
    \end{subtable}

  \begin{subtable}{\textwidth}
    \centering
    \resizebox{\textwidth}{!}{
    \small
      \begin{tabular}{ccccc}
        \toprule
        \textbf{Method} & Temporal Relational & Timepoint Indexed & Multifaceted Integrative & \textbf{Average} \\
        \midrule
        ReSimplifyIt (Ours) & 66.8 & 78.5 & 72.8 & 72.7 \\
        \bottomrule
      \end{tabular}
    }
    \caption{Results on query rewriting}
    \label{tab:emc_main_reduced_text}
  \end{subtable}

  \caption{Stage-1 evaluation results on EMCompress dataset.}
  \label{tab:stage1_eval}
\end{table*}

\begin{table*}[t]
\centering
\setlength{\tabcolsep}{2.5pt} 
\resizebox{\linewidth}{!}{
\small
\begin{tabular}{l|c|cc|cc|cc|cc|cc|cc|cc}
\toprule
\multirow{2}{*}{\textbf{Method}} & \multirow{2}{*}{\textbf{Size}} & \multicolumn{2}{c|}{\textbf{ActivityNetQA}} & \multicolumn{2}{c|}{\textbf{EMCompress}} & \multicolumn{2}{c|}{\textbf{EgoSchema}} & \multicolumn{2}{c|}{\textbf{LVBench}} & \multicolumn{2}{c|}{\textbf{MLVU}} & \multicolumn{2}{c|}{\textbf{Video-MME}} & \multicolumn{2}{c}{\textbf{NExT-OE}} \\
& & w/emc & w/o & w/emc & w/o & w/emc & w/o & w/emc & w/o & w/emc & w/o & w/emc & w/o & w/emc & w/o \\
\midrule
Video-ChatGPT \citep{maaz-etal-2024-video} & 7B & \textbf{58.5} & 50.5 & \textbf{38.9} & 28.91 & \textbf{29.2} & 23.0 & \textbf{23.5} & 22.9 & \textbf{22.5} & 18.6 & \textbf{31.1} & 29.5 & \textbf{49.8} & 41.5 \\
Video-LLaVA \citep{lin2023video} & 7B & \textbf{61.2} & 52.1 & \textbf{43.2} & 32.4 & \textbf{43.8} & 40.0 & \textbf{26.5} & 23.2 & \textbf{31.9} & 27.5 & \textbf{43.4} & 39.1 & \textbf{48.0} & 43.3 \\
ChatUniVi \citep{chatunivi} & 7B & \textbf{63.2} & 52.0 & \textbf{56.5} & 47.4 & -- & -- & -- & -- & -- & -- & -- & -- & \textbf{34.5} & 25.9 \\
LLaVA-NExT \citep{liu2024llavanext} & 7B & \textbf{67.8} & 62.5 & \textbf{58.2} & 48.6 & \textbf{38.6} & 38.2 & \textbf{31.4} & 23.8 & \textbf{34.2} & 28.3 & \textbf{43.8} & 40.3 & \textbf{52.3} & 43.7 \\
InternVL3.5 \citep{InternVL3} & 8B & \textbf{59.9} & 57.2 & \textbf{51.5} & 45.7 & \textbf{67.2} & 61.8 & \textbf{46.3} & 39.6 & \textbf{46.8} & 45.1 & \textbf{60.4} & 56.9 & \textbf{59.1} & 53.9 \\
InternVL3.5 \citep{InternVL3} & 14B & \textbf{60.1} & 58.9 & \textbf{52.1} & 48.8 & \textbf{70.6} & 67.6 & \textbf{47.8} & 42.9 & \textbf{46.5} & 45.2 & \textbf{62.2} & 61.1 & \textbf{58.2} & 55.6 \\
Qwen2.5-VL \citep{qwen2.5vl} & 3B & \textbf{57.6} & 55.8 & \textbf{47.9} & 40.3 & \textbf{61.8} & 57.2 & \textbf{44.2} & 36.3 & \textbf{45.3} & 41.3 & \textbf{59.0} & 57.7 & \textbf{54.8} & 51.6 \\
Qwen2.5-VL \citep{qwen2.5vl} & 7B & \textbf{58.0} & 55.2 & \textbf{48.0} & 41.1 & \textbf{68.2} & 66.4 & \textbf{43.0} & 34.8 & \textbf{43.1} & 37.6 & \textbf{60.0} & 58.5 & \textbf{55.2} & 53.4 \\
Qwen3-VL \citep{qwen3vl} & 4B & \textbf{59.9} & 57.7 & \textbf{48.0} & 45.6 & \textbf{71.2} & 70.2 & \textbf{43.2} & 37.5 & \textbf{49.0} & 42.8 & 60.5 & \textbf{60.6} & \textbf{61.1} & 58.2 \\
Qwen3-VL \citep{qwen3vl} & 32B & \textbf{60.9} & 60.5 & \textbf{47.8} & 45.2 & \textbf{72.8} & 72.6 & -- & -- & \textbf{44.9} & 38.7 & \textbf{64.0} & 61.2 & \textbf{61.4} & 58.9 \\
LLaVA-OneVision \citep{llavaonevision} & 4B & -- & -- & -- & -- & \textbf{18.6} & 18.2 & -- & -- & \textbf{13.5} & 5.5 & -- & -- & -- & -- \\
LLaVA-OneVision \citep{llavaonevision} & 8B & \textbf{32.4} & 27.5 & -- & -- & \textbf{38.4} & 32.5 & -- & -- & \textbf{14.1} & 12.2 & 51.1 & \textbf{51.3} & -- & -- \\
\midrule
VideoAgent \citep{VideoAgent} & - & \textcolor{gray}{\textbf{61.5}} & \textcolor{gray}{60.2} & \textcolor{gray}{\textbf{53.9}} & \textcolor{gray}{38.4} & -- & -- & -- & -- & -- & -- & -- & -- & \textcolor{gray}{47.8} & \textcolor{gray}{\textbf{49.6}} \\
VideoTree \citep{wang2024videotreeadaptivetreebasedvideo} & - & \textcolor{gray}{\textbf{63.6}} & \textcolor{gray}{59.0} & \textcolor{gray}{\textbf{69.4}} & \textcolor{gray}{57.1} & -- & -- & -- & -- & -- & -- & -- & -- & \textcolor{gray}{\textbf{61.9}} & \textcolor{gray}{57.7} \\
\midrule
GPT-4o \citep{gpt4o} & - & \textbf{75.65} & 72.2 & \textbf{73.84} & 63.59 & \textbf{72.38} & 71.17 & \textbf{46.22} & 35.04 & \textbf{48.92} & 44.73 & \textbf{66.91} & 61.63 & \textbf{62.25} & 54.9 \\
GPT-4.1-mini \citep{gpt4} & - & \textbf{77.07} & 75.19 & \textbf{76.04} & 68.74 & 71.12 & \textbf{72.02} & \textbf{46.35} & 34.43 & \textbf{50.67} & 39.90 & \textbf{62.45} & 61.20 & \textbf{68.93} & 57.45 \\
GPT-4-turbo \citep{gpt4} & - & \textbf{77.1} & 74.34 & \textbf{79.39} & 70.61 & \textbf{69.20} & 66.60 & -- & -- & -- & -- & -- & -- & \textbf{70.35} & 61.4 \\
\bottomrule
\end{tabular}
}
\caption{Evaluation results of EMC-guided inference. \textbf{Bold} values indicate the better-performing result for each baseline, comparing the \textit{\textbf{with EMC}} \textit{v.s.} \textit{\textbf{without EMC}} configurations on each benchmark.}
\label{stage2_eval}
\vspace{-0.3em}
\end{table*}

\begin{table*}[t]
  \centering
  \resizebox{\linewidth}{!}{
  \small
  \begin{tabular}{cc|cc|cc|cc|cc}
    \toprule
    \multicolumn{2}{c|}{\textbf{Method}} & \multicolumn{2}{c|}{\textbf{ActivityNetQA}} & \multicolumn{2}{c|}{\textbf{EMCompress}} & \multicolumn{2}{c|}{\textbf{NExT-QA}} & \multicolumn{2}{c}{\textbf{NExT-OE}} \\
    & & w/emc & w/o & w/emc & w/o & w/emc & w/o & w/emc & w/o \\
    \midrule
    \multirow{3}{*}{Video-ChatGPT} 
      & w/o emc        & 57.4 & 49.8 & 45.1 & 35.6 & 48.6 & 41.3 & 46.2 & 40.0 \\
      & w/ emc (gt)    & 62.7 & 54.4 & 49.8 & 38.9 & 51.0 & 45.7 & 49.2 & 43.5 \\
      & w/ emc (pred)  & 60.0 & 51.3 & 48.2 & 37.1 & 49.6 & 44.3 & 48.7 & 41.9 \\
    \midrule
    \multirow{3}{*}{Video-LLaVA} 
      & w/o emc        & 53.9 & 45.1 & 46.3 & 35.9 & 50.2 & 47.8 & 47.0 & 39.1 \\
      & w/ emc (gt)    & 57.0 & 48.9 & 52.6 & 41.6 & 55.7 & 50.8 & 52.1 & 44.9 \\
      & w/ emc (pred)  & 54.2 & 46.9 & 48.1 & 37.4 & 51.0 & 49.3 & 50.7 & 41.8 \\
    \midrule
    \multirow{3}{*}{ChatUniVi} 
      & w/o emc        & 63.2 & 50.0 & 62.4 & 57.4 & --   & --   & 29.0 & 22.5 \\
      & w/ emc (gt)    & 65.6 & 54.4 & 67.8 & 63.8 & --   & --   & 34.3 & 26.0 \\
      & w/ emc (pred)  & 64.9 & 52.6 & 62.8 & 61.6 & --   & --   & 32.1 & 24.0 \\
    \bottomrule
  \end{tabular}
  }
  \caption{Evaluation results of EMC-guided training by downstream inference. On the rows, \textit{w/o emc}, \textit{w/emc (gt)}, and \textit{w/emc (pred)} indicate training with vanilla EMCompress dataset, training with ground truth EMC labels, and training with the predicted EMC results, respectively; on the columns, \textit{w/o emc} and \textit{w/emc} indicate the inference mode.
  }
  \label{stage3_eval}
\end{table*}

\addressed{\uky{Results are subject to small changes. }}

\subsection{EMCompress}
\label{yc2emc}
As EMC is a new task, few benchmarks are capable of performing the evaluation. While some studies \citep{tvqa, groundedmultihopvideoqalongform, tvqa+} have explored grounded VideoQA which may be mistakenly seen as equivalent, they mainly focus on improving visual evidence. See Appendix \ref{appendix: related_work} for details. The synchronous update of video and query inputs in EMC naturally resolves any potential semantic mismatch, maximally enhancing generalizability and adaptability of our work.

To bridge this gap, we introduce the EMCompress benchmark, which provides both EMC and standard VideoQA labels. Built upon the YouCookII dataset \citep{ZhXuCoAAAI18}, it comprises 2,754 datapoints split into training, validation, and test sets (roughly 7:1:2). This dual-task benchmark enables unified evaluation of both EMC and VideoQA. Refer to Appendix~\ref{yc2emc_appendix} for more details.

\subsection{Datasets and Benchmarks}
We conduct the evaluation from three aspects.

Firstly, we evaluate the quality of the EMC process on \textbf{EMCompress} (Section~\ref{yc2emc}). Next, we examine the impact of EMC on downstream VideoQA performance across the following benchmarks: \textbf{EMCompress}, which supports evaluation of both the EMC process and VideoQA performance; \textbf{ActivityNet-QA}~\citep{yu2019activityqa}, \textbf{NExT-OE}~\citep{xiao2021next}, \textbf{Video-MME}~\citep{videomme}, \textbf{MLVU}~\citep{mlvu}, \textbf{LVBench}~\citep{lvbench}, and \textbf{EgoSchema}~\citep{egoschema}. Lastly, we investigate the role of EMC in video instruction tuning by fine-tuning Video-LLM baselines on \textbf{EMCompress}, and comparing their downstream performance across various benchmarks. For \textbf{NExT-QA}~\citep{xiao2021next}, we conduct open-ended generation during inference and map predictions to MCQ format using GPT-3.5, ensuring consistent evaluation across baselines and benchmarks (see the Appendix for details).

\subsection{Implementation Details}
We provide implementation details in Appendix \ref{appendix:imp}.

\subsection{Baselines}

\paragraph{Endomorphic Multimodal Compression}
 Considering the lack of baselines of the new EMC task, we adopt two baselines from the training-free temporal localization task which are considered to be robust and generalizable to unseen datasets, to make a side-by-side comparison of the video output alone of EMC. Specifically, we adopt VTG-GPT \citep{vtggpt}, a proposal-based method which made one of the first attempts to training-free video temporal grounding, and \citet{tfvtg}'s work which comprehend candidate proposals based on both static and dynamic matching scores. For the query output of the EMC process, we report open-ended evaluation result of our proposed framework.

\addressed{\jiateng{I think the Video-QA part should be put into the method section. Open a new subsection called: EMC Guided Training for Video LLMs and discuss our appraoch to train the model.}\uky{done}}

\paragraph{EMC-guided inference}
Integrating EMC as a front-end module into existing VideoQA agents yields a novel VideoQA framework. We adopt several Video-LLMs---Video-ChatGPT~\citep{maaz-etal-2024-video}, Video-LLaVA~\citep{lin2023video}, ChatUniVi~\citep{chatunivi}, LLaVA-NeXT~\citep{liu2024llavanext}, InternVL3.5~\citep{InternVL3}, Qwen2.5-VL~\citep{qwen2.5vl}, Qwen3-VL~\citep{qwen3vl}, LLaVA-OneVision~\citep{llavaonevision}---as representative baselines. We also examine LLM-assisted frameworks such as VideoTree~\citep{wang2024videotreeadaptivetreebasedvideo} and VideoAgent~\citep{VideoAgent}, which, unlike the static encoding approaches of Video-LLMs, dynamically extract video frames based on the textual query.

\paragraph{EMC-guided training}
Following Section \ref{subsection:emc-guided-training}, we performed video instruction tuning on Video-ChatGPT \citep{maaz-etal-2024-video}, Video-LLaVA \citep{lin2023video}, and ChatUniVi \citep{chatunivi} on EMCompress. We kept the LLM backbone frozen, and only tuned their multimodal projectors.
\section{Evaluation Results}

The EMC task addresses a core bottleneck in VideoQA: enabling models to focus on semantically aligned moments in long videos by abstracting task-level semantics and guiding modality-aware information filtering, which in turn enhances multi-modal alignment. We validate EMC's importance through two complementary tests: (1) ground-truth EMC significantly boosts performance (e.g., +7.3\% in Table \ref{stage3_eval}), confirming it as a key supervision signal; (2) current methods fail to solve EMC effectively (Table \ref{tab:stage1_eval}), motivating our ReSimplifyIt as a plug-and-play approach.

\paragraph{Endomorphic Multimodal Compression} Our ReSimplifyIt framework outperforms both baselines significantly, on every metric and subset of our EMCompress benchmark. On the average performance of the whole test set of EMCompress, ReSimplifyIt achieved an mIoU score higher than 300\% of the baselines' performance. For query rewriting, our proposed framework also achieved notably good performance. Refer to Table \ref{tab:stage1_eval} and Table \ref{tab:stage1_eval_appendix} for results. We provide the prompt in Appendix \ref{prompts}.

\subparagraph{Ablation Study of ReSimplifyIt Framework}
We conduct an ablation study on the ReSimplifyIt framework to evaluate the effectiveness of its design. More details are provided in Appendix \ref{ablations}.

\paragraph{EMC guided inference}
As shown in Table \ref{stage2_eval} and Table \ref{stage2_eval_appendix}, nearly all baselines---ranging from open source models of different sizes and architectures to proprietary models---saw solid absolute performance gain, suggesting that Video-LLMs largely suffer from superfluous cognition noise. The \textbf{consistent performance gains across all benchmarks} underscore the universality of this general reasoning load optimization problem across diverse tasks and scenarios. In contrast, both LLM-assisted reasoning frameworks saw smaller gains. We hypothesize that the design and implementation of these frameworks have intrinsically incorporated the EMC process, by leveraging strong reasoning ability of external LLMs or multi-turn interaction.

\paragraph{EMC guided training} All checkpoints trained on ground truth EMC labels achieved stable performance gain over their counterparts trained on the vanilla EMCompress benchmark, while the performance gain of the checkpoints trained on the predicted EMC output of ReSimplifyIt was weaker. We argue that EMC-guided training benefits Video-LLMs better, particularly when faced with untrimmed videos and complex text instructions requiring multi-hop reasoning. Refer to Table \ref{stage3_eval} for further details.

\section{Efficiency Analysis}
\label{sec:efficiency}

We now quantify the practical overhead of the EMC compression loop against the savings it yields downstream. For comparability under variable conditions, we report \textit{hardware-agnostic proxies}---LLM/tool/caption call counts, output tokens, duration reduction, and compression success rate---together with closed-loop metrics under realistic frame budgets.

\paragraph{Cost drivers and compression effectiveness.}
For each sample we record the number of external LLM turns, tool invocations, total captions, and output tokens, together with the duration ratios $\mathrm{DurAll} = \mathbb{E}[|v|/|V|]$ and $\mathrm{DurScrn}$ (restricted to successfully compressed samples), and $\mathrm{Compress\%}$, the fraction of samples for which $\mathcal{F}_{\mathrm{EMC}}(V,Q) \ne (V,Q)$. See Appendix~\ref{appendix:efficiency_full} (Tables~\ref{tab:cost_drivers_simple_app}, \ref{tab:cost_drivers_full_app}, and~\ref{tab:cost_full_breakdown}) for per-dataset values and the full per-source caption breakdown. ReSimplifyIt-simple averages only ${\sim}22$ captions, $3$--$10$ LLM calls, and $300$--$1{,}500$ output tokens per sample, succeeding on $94.4\%$--$100\%$ of samples across all seven benchmarks. The full multi-agent ReSimplifyIt uses more resources due to its iterative refinement, with roughly half of its captions pre-loaded in agent prompts and the rest fetched via the Viewer's tools. Both variants reduce videos to a small fraction of their original duration on successfully compressed samples---down to $1.9\%$ (simple) and $4.8\%$ (full) on LVBench---with compression strongest on the longest benchmarks.

\paragraph{Closing the loop under fixed frame budgets.}
A naive comparison via the count of downstream frames is misleading: a fixed budget of $K$ frames sampled uniformly over a long video is extremely sparse and may under-cover the evidence, while the same $K$ frames concentrated on the compressed segment yield much denser coverage. To make this precise, we define
\begin{align*}
\mathrm{DensAmp} &= \mathbb{E}\!\left[\frac{1}{\mathrm{DurRatio}}\right]\!, \\
\mathrm{EquivFr}(K) &= K \cdot \mathrm{DensAmp}, \\
\mathrm{CostRatio}(K) &= \frac{K + \mathrm{TotalCap}}{\mathrm{EquivFr}(K)}.
\end{align*}
Here $\mathrm{DensAmp}$ captures how much denser downstream sampling becomes after compression: $K$ post-compression frames are equivalent to $\mathrm{EquivFr}(K)$ frames uniformly sampled over the original video. $\mathrm{CostRatio}(K)$ then expresses the total visual sampling cost ($K + \mathrm{TotalCap}$) as a fraction of the dense-sampling cost needed to reach the same evidence density. A lower $\mathrm{CostRatio}$ therefore means EMC achieves the same effective evidence density at a smaller fraction of the dense-sampling cost: $\mathrm{CostRatio}{=}0.1$ means the no-EMC baseline would incur $10\times$ the frame-sampling cost \emph{to reach the same evidence density}. We also report the output-token cost of each $1\%$ reduction in video duration,
\[
\mathrm{OutTok/1\%Red} = \frac{\mathbb{E}[\#\mathrm{OutTok}]}{(1 - \mathbb{E}[\mathrm{DurScrn}])\cdot 100}.
\]

\begin{table}[!t]
  \centering
  \scriptsize
  \setlength{\tabcolsep}{3pt}
  \begin{subtable}{\columnwidth}
    \centering
    \resizebox{\columnwidth}{!}{%
    \begin{tabular}{l|cc|cccc}
      \toprule
      \textbf{Dataset} & \textbf{DensAmp} & \textbf{OT/1\%} & \multicolumn{4}{c}{\textbf{CostRatio}$(K)$} \\
      & & & $K{=}8$ & $K{=}16$ & $K{=}32$ & $K{=}100$ \\
      \midrule
      ActivityNet-QA  & 25.0$\times$  & 3.8  & 0.15 & 0.09 & 0.07 & 0.05 \\
      EMCompress   & 29.7$\times$  & 3.9  & 0.13 & 0.08 & 0.06 & 0.04 \\
      NExT-OE           & 20.7$\times$  & 5.2  & 0.19 & 0.12 & 0.08 & 0.06 \\
      EgoSchema       & 6.3$\times$   & 6.0  & 0.59 & 0.38 & 0.27 & 0.19 \\
      LVBench         & 320.7$\times$ & 15.1 & 0.01 & 0.01 & 0.01 & 0.00 \\
      MLVU            & 70.6$\times$  & 10.8 & 0.06 & 0.04 & 0.02 & 0.02 \\
      Video-MME       & 51.2$\times$  & 8.8  & 0.08 & 0.05 & 0.03 & 0.02 \\
      \bottomrule
    \end{tabular}%
    }
    \caption{ReSimplifyIt-simple}
    \label{tab:costratio_simple}
  \end{subtable}

  \vspace{0.3em}
  \begin{subtable}{\columnwidth}
    \centering
    \resizebox{\columnwidth}{!}{%
    \begin{tabular}{l|cc|cccc}
      \toprule
      \textbf{Dataset} & \textbf{DensAmp} & \textbf{OT/1\%} & \multicolumn{4}{c}{\textbf{CostRatio}$(K)$} \\
      & & & $K{=}8$ & $K{=}16$ & $K{=}32$ & $K{=}100$ \\
      \midrule
      ActivityNet-QA  & 65.2$\times$  & 39.6 & 0.18 & 0.10 & 0.06 & 0.03 \\
      EMCompress   & 14.7$\times$  & 25.5 & 0.70 & 0.38 & 0.23 & 0.12 \\
      NExT-OE           & 12.3$\times$  & 34.9 & 0.83 & 0.45 & 0.27 & 0.14 \\
      EgoSchema       & 25.8$\times$  & 77.3 & 0.67 & 0.36 & 0.20 & 0.09 \\
      LVBench         & 852.8$\times$ & 38.5 & 0.02 & 0.01 & 0.01 & 0.00 \\
      MLVU            & 308.5$\times$ & 48.9 & 0.06 & 0.03 & 0.02 & 0.01 \\
      Video-MME       & 114.5$\times$ & 43.2 & 0.12 & 0.07 & 0.04 & 0.02 \\
      \bottomrule
    \end{tabular}%
    }
    \caption{ReSimplifyIt (full).}
    \label{tab:costratio_full}
  \end{subtable}
  \caption{Density amplification (DensAmp), output-token cost per $1\%$ of video-length reduction (OT/1\%), and visual CostRatio under four downstream frame budgets $K$. Statistics are averaged over successfully compressed samples.}
  \label{tab:costratio}
\end{table}

\paragraph{Density amplification and end-to-end cost.}
Table~\ref{tab:costratio} quantifies how the compression overhead is repaid by concentrated downstream sampling. ReSimplifyIt-simple achieves $6.3\times$--$320.7\times$ density amplification; on LVBench, $8$ frames sampled from the compressed segment match the temporal density of $2{,}566$ frames sampled uniformly over the full video. The full ReSimplifyIt reaches $12.3\times$--$852.8\times$ on successfully compressed samples, with output-token costs of $3.8$--$15.1$ tokens per $1\%$ duration reduction (simple) and $25.5$--$77.3$ (full). At $K{=}8$, the simple variant's $\mathrm{CostRatio}\leq 0.19$ on $6$ of $7$ datasets (LVBench reaching $0.01$, a ${\sim}100\times$ reduction); the full variant reports $\mathrm{CostRatio}\in[0.02, 0.18]$ on the long-video benchmarks where it is most effective. As $K$ grows to $16, 32, 100$, $\mathrm{CostRatio}$ decreases monotonically across all datasets, indicating that EMC becomes \emph{increasingly} cost-effective as downstream Video-LLMs adopt denser frame sampling. Combined with the accuracy gains of the previous section, EMC is a practically deployable front-end whose three instantiations (Appendix~\ref{ablations}) further offer an explicit performance--cost spectrum.

\section{Related Work} 
We provide more details in appendix \ref{appendix: related_work}.
\paragraph{Video-LLMs for VideoQA}
Video-LLMs have spurred a wave of models aimed at enhancing video understanding~\citep{lin2023video, Ma_2024_CVPR, 10658165, Li2023LLaMAVIDAI, liu2024llavanext, Xu2024PLLaVAP}, while the sparsity and query-invariant nature of their encoding limits efficacy in capturing fine-grained spatial-temporal details.

\paragraph{LLM-assisted Agentic Reasoning for VideoQA}
Some other work proposing robust VideoQA baselines opt to explore pure-text LLM assisted frameworks or multi-agent systems for VideoQA \citep{wang2024videotreeadaptivetreebasedvideo, shang-etal-2024-traveler, VideoAgent}. These methods adopt LLM-based methods to serve as a scheduler, which implicitly fulfills the EMC objective to a significant extent.

\section{Conclusion}
\vspace{-0.4em}
We introduce Endomorphic Multimodal Compression (EMC), a cognitively inspired task that reconstructs each VideoQA instance $(V, Q)$ into an answer-preserving compact pair $(v, q)$ to purify cognition load. We realize EMC via ReSimplifyIt, a plug-and-play, model-agnostic multi-agent framework, and release EMCompress to benchmark compression-centric reasoning. Across models and datasets, EMC consistently strengthens both training-time alignment and inference-time robustness, yielding notable accuracy gains and exposing bottlenecks via compact, auditable rationales, positioning information compression as a key direction for advancing video-language understanding.

\section*{Acknowledgments}

This research is based upon work supported by U.S. DARPA ECOLE Program No. \#HR00112390060. The views and conclusions contained herein are those of the authors and should not be interpreted as necessarily representing the official policies, either expressed or implied, of DARPA, or the U.S. Government. The U.S. Government is authorized to reproduce and distribute reprints for governmental purposes notwithstanding any copyright annotation therein.

\section*{Limitations}

While endomorphic multimodal compression brings considerable performance gains, limitations remain and leave room for future work.
Firstly, endomorphic multimodal compression is applied only on the temporal axis and cannot filter redundant spatial visual information. We regard spatial visual compression as a natural next frontier, complementary to the temporal compression studied here and opening an orthogonal axis of the broader information compression agenda.
Secondly, while our plug-and-play framework adapts to any VideoQA agent, its end-to-end counterpart---Video-LLMs with inter-frame reasoning or query-adaptive frame sampling built into the encoder---is left unexplored. Both external adapters and end-to-end integration can advance information compression.

\bibliography{custom}

\begin{thebibliography}{64}
\providecommand{\natexlab}[1]{#1}

\bibitem[{Alemi et~al.(2017)Alemi, Fischer, Dillon, and Murphy}]{alemi2017deep}
Alexander~A. Alemi, Ian Fischer, Joshua~V. Dillon, and Kevin Murphy. 2017.
\newblock Deep variational information bottleneck.
\newblock In \emph{International Conference on Learning Representations
  (ICLR)}.

\bibitem[{Bai et~al.(2025{\natexlab{a}})Bai, Cai, Chen, Chen, Chen, Cheng,
  Deng, Ding, Gao, Ge, Ge, Guo, Huang, Huang, Huang, Hui, Jiang, Li, Li, Li,
  Li, Lin, Lin, Liu, Liu, Liu, Liu, Liu, Liu, Lu, Luo, Lv, Men, Meng, Ren, Ren,
  Song, Sun, Tang, Tu, Wan, Wang, Wang, Wang, Wang, Xie, Xu, Xu, Xu, Yang,
  Yang, Yang, Yang, Yu, Zhang, Zhang, Zhang, Zheng, Zhong, Zhou, Zhou, Zhou,
  Zhu, and Zhu}]{qwen3vl}
Shuai Bai, Yuxuan Cai, Ruizhe Chen, Keqin Chen, Xionghui Chen, Zesen Cheng,
  Lianghao Deng, Wei Ding, Chang Gao, Chunjiang Ge, Wenbin Ge, Zhifang Guo,
  Qidong Huang, Jie Huang, Fei Huang, Binyuan Hui, Shutong Jiang, Zhaohai Li,
  Mingsheng Li, and 45 others. 2025{\natexlab{a}}.
\newblock Qwen3-vl technical report.
\newblock \emph{arXiv preprint arXiv:2511.21631}.

\bibitem[{Bai et~al.(2025{\natexlab{b}})Bai, Chen, Liu, Wang, Ge, Song, Dang,
  Wang, Wang, Tang, Zhong, Zhu, Yang, Li, Wan, Wang, Ding, Fu, Xu, Ye, Zhang,
  Xie, Cheng, Zhang, Yang, Xu, and Lin}]{qwen2.5vl}
Shuai Bai, Keqin Chen, Xuejing Liu, Jialin Wang, Wenbin Ge, Sibo Song, Kai
  Dang, Peng Wang, Shijie Wang, Jun Tang, Humen Zhong, Yuanzhi Zhu, Mingkun
  Yang, Zhaohai Li, Jianqiang Wan, Pengfei Wang, Wei Ding, Zheren Fu, Yiheng
  Xu, and 8 others. 2025{\natexlab{b}}.
\newblock \href {https://arxiv.org/abs/2502.13923} {Qwen2.5-vl technical
  report}.
\newblock \emph{Preprint}, arXiv:2502.13923.

\bibitem[{Bertasius et~al.(2021)Bertasius, Wang, and
  Torresani}]{gberta_2021_ICML}
Gedas Bertasius, Heng Wang, and Lorenzo Torresani. 2021.
\newblock Is space-time attention all you need for video understanding?
\newblock In \emph{Proceedings of the International Conference on Machine
  Learning (ICML)}.

\bibitem[{Besta et~al.(2024)Besta, Blach, Kubicek, Gerstenberger, Podstawski,
  Gianinazzi, Gajda, Lehmann, Niewiadomski, Nyczyk, and
  Hoefler}]{besta2024graph}
Maciej Besta, Nils Blach, Ales Kubicek, Robert Gerstenberger, Michal
  Podstawski, Lukas Gianinazzi, Joanna Gajda, Tomasz Lehmann, Hubert
  Niewiadomski, Piotr Nyczyk, and Torsten Hoefler. 2024.
\newblock Graph of thoughts: Solving elaborate problems with large language
  models.
\newblock In \emph{Proceedings of the AAAI Conference on Artificial
  Intelligence}, volume~38, pages 17682--17690.

\bibitem[{Brooks(1991)}]{brooks1991intelligence}
Rodney~A Brooks. 1991.
\newblock Intelligence without representation.
\newblock \emph{Artificial intelligence}, 47(1-3):139--159.

\bibitem[{Chen et~al.(2024{\natexlab{a}})Chen, Liu, Huang, He, Pei, Xu, Wang,
  Lu, and Wang}]{cgbench}
Guo Chen, Yicheng Liu, Yifei Huang, Yuping He, Baoqi Pei, Jilan Xu, Yali Wang,
  Tong Lu, and Limin Wang. 2024{\natexlab{a}}.
\newblock \href {https://arxiv.org/abs/2412.12075} {Cg-bench: Clue-grounded
  question answering benchmark for long video understanding}.
\newblock \emph{Preprint}, arXiv:2412.12075.

\bibitem[{Chen et~al.(2023)Chen, Zheng, Wang, Xu, Huang, Pan, Wang, Wang, Qiao,
  Lu, and Wang}]{chen2023video-LLMmodelingvideosequence}
Guo Chen, Yin-Dong Zheng, Jiahao Wang, Jilan Xu, Yifei Huang, Junting Pan,
  Yi~Wang, Yali Wang, Yu~Qiao, Tong Lu, and Limin Wang. 2023.
\newblock \href {https://arxiv.org/abs/2305.13292} {Videollm: Modeling video
  sequence with large language models}.
\newblock \emph{Preprint}, arXiv:2305.13292.

\bibitem[{Chen et~al.(2024{\natexlab{b}})Chen, Di, and
  Xie}]{groundedmultihopvideoqalongform}
Qirui Chen, Shangzhe Di, and Weidi Xie. 2024{\natexlab{b}}.
\newblock \href {https://arxiv.org/abs/2408.14469} {Grounded multi-hop videoqa
  in long-form egocentric videos}.
\newblock \emph{Preprint}, arXiv:2408.14469.

\bibitem[{Chen and Jiang(2019)}]{10.1609/aaai.v33i01.33018199}
Shaoxiang Chen and Yu-Gang Jiang. 2019.
\newblock \href {https://doi.org/10.1609/aaai.v33i01.33018199} {Semantic
  proposal for activity localization in videos via sentence query}.
\newblock In \emph{Proceedings of the Thirty-Third AAAI Conference on
  Artificial Intelligence and Thirty-First Innovative Applications of
  Artificial Intelligence Conference and Ninth AAAI Symposium on Educational
  Advances in Artificial Intelligence}, AAAI'19/IAAI'19/EAAI'19. AAAI Press.

\bibitem[{Cover and Thomas(2006)}]{cover2006elements}
Thomas~M. Cover and Joy~A. Thomas. 2006.
\newblock \emph{Elements of Information Theory}, 2nd edition.
\newblock Wiley-Interscience.

\bibitem[{Fu et~al.(2024)Fu, Dai, Luo, Li, Ren, Zhang, Wang, Zhou, Shen, Zhang
  et~al.}]{videomme}
Chaoyou Fu, Yuhan Dai, Yondong Luo, Lei Li, Shuhuai Ren, Renrui Zhang, Zihan
  Wang, Chenyu Zhou, Yunhang Shen, Mengdan Zhang, and 1 others. 2024.
\newblock Video-mme: The first-ever comprehensive evaluation benchmark of
  multi-modal llms in video analysis.
\newblock \emph{arXiv preprint arXiv:2405.21075}.

\bibitem[{Gao et~al.(2017)Gao, Sun, Yang, and Nevatia}]{8237825}
Jiyang Gao, Chen Sun, Zhenheng Yang, and Ram Nevatia. 2017.
\newblock \href {https://doi.org/10.1109/ICCV.2017.563} {Tall: Temporal
  activity localization via language query}.
\newblock In \emph{2017 IEEE International Conference on Computer Vision
  (ICCV)}, pages 5277--5285.

\bibitem[{Hendricks et~al.(2017)Hendricks, Wang, Shechtman, Sivic, Darrell, and
  Russell}]{8237880}
Lisa~Anne Hendricks, Oliver Wang, Eli Shechtman, Josef Sivic, Trevor Darrell,
  and Bryan Russell. 2017.
\newblock \href {https://doi.org/10.1109/ICCV.2017.618} {Localizing moments in
  video with natural language}.
\newblock In \emph{2017 IEEE International Conference on Computer Vision
  (ICCV)}, pages 5804--5813.

\bibitem[{Hochstein and Ahissar(2002)}]{topdownfeedback}
Shaul Hochstein and Merav Ahissar. 2002.
\newblock \href {https://doi.org/10.1016/S0896-6273(02)01091-7} {View from the
  top: Hierarchies and reverse hierarchies in the visual system}.
\newblock \emph{Neuron}, 36(5):791--804.

\bibitem[{Jin et~al.(2023)Jin, Takanobu, Zhang, Cao, and Yuan}]{chatunivi}
Peng Jin, Ryuichi Takanobu, Caiwan Zhang, Xiaochun Cao, and Li~Yuan. 2023.
\newblock Chat-univi: Unified visual representation empowers large language
  models with image and video understanding.
\newblock \emph{arXiv preprint arXiv:2311.08046}.

\bibitem[{Le~Gall(1991)}]{mpeg}
Didier Le~Gall. 1991.
\newblock \href {https://doi.org/10.1145/103085.103090} {Mpeg: a video
  compression standard for multimedia applications}.
\newblock \emph{Commun. ACM}, 34(4):46–58.

\bibitem[{Lei et~al.(2018)Lei, Yu, Bansal, and Berg}]{tvqa}
Jie Lei, Licheng Yu, Mohit Bansal, and Tamara Berg. 2018.
\newblock \href {https://doi.org/10.18653/v1/D18-1167} {{TVQA}: Localized,
  compositional video question answering}.
\newblock In \emph{Proceedings of the 2018 Conference on Empirical Methods in
  Natural Language Processing}, pages 1369--1379, Brussels, Belgium.
  Association for Computational Linguistics.

\bibitem[{Lei et~al.(2020)Lei, Yu, Berg, and Bansal}]{tvqa+}
Jie Lei, Licheng Yu, Tamara Berg, and Mohit Bansal. 2020.
\newblock \href {https://doi.org/10.18653/v1/2020.acl-main.730} {Tvqa+:
  Spatio-temporal grounding for video question answering}.
\newblock In \emph{Proceedings of the 58th Annual Meeting of the Association
  for Computational Linguistics}, pages 8211--8225, Online. Association for
  Computational Linguistics.

\bibitem[{Li et~al.(2024{\natexlab{a}})Li, Zhang, Guo, Zhang, Li, Zhang, Zhang,
  Li, Liu, and Li}]{llavaonevision}
Bo~Li, Yuanhan Zhang, Dong Guo, Renrui Zhang, Feng Li, Hao Zhang, Kaichen
  Zhang, Yanwei Li, Ziwei Liu, and Chunyuan Li. 2024{\natexlab{a}}.
\newblock Llava-onevision: Easy visual task transfer.
\newblock \emph{arXiv preprint arXiv:2408.03326}.

\bibitem[{Li et~al.(2023)Li, He, Wang, Li, Wang, Luo, Wang, Wang, and
  Qiao}]{2023videochat}
Kunchang Li, Yinan He, Yi~Wang, Yizhuo Li, Wenhai Wang, Ping Luo, Yali Wang,
  Limin Wang, and Yu~Qiao. 2023.
\newblock Videochat: Chat-centric video understanding.
\newblock \emph{arXiv preprint arXiv:2305.06355}.

\bibitem[{Li et~al.(2024{\natexlab{b}})Li, Wang, He, Li, Wang, Liu, Wang, Xu,
  Chen, Lou, Wang, and Qiao}]{10658165}
Kunchang Li, Yali Wang, Yinan He, Yizhuo Li, Yi~Wang, Yi~Liu, Zun Wang, Jilan
  Xu, Guo Chen, Ping Lou, Limin Wang, and Yu~Qiao. 2024{\natexlab{b}}.
\newblock \href {https://doi.org/10.1109/CVPR52733.2024.02095} {Mvbench: A
  comprehensive multi-modal video understanding benchmark}.
\newblock In \emph{2024 IEEE/CVF Conference on Computer Vision and Pattern
  Recognition (CVPR)}, pages 22195--22206.

\bibitem[{Li et~al.(2024{\natexlab{c}})Li, Wang, and Jia}]{Li2023LLaMAVIDAI}
Yanwei Li, Chengyao Wang, and Jiaya Jia. 2024{\natexlab{c}}.
\newblock \href {https://api.semanticscholar.org/CorpusID:265466723}
  {Llama-vid: An image is worth 2 tokens in large language models}.
\newblock In \emph{European Conference on Computer Vision}.

\bibitem[{Lin et~al.(2023)Lin, Zhu, Ye, Ning, Jin, and Yuan}]{lin2023video}
Bin Lin, Bin Zhu, Yang Ye, Munan Ning, Peng Jin, and Li~Yuan. 2023.
\newblock Video-llava: Learning united visual representation by alignment
  before projection.
\newblock \emph{arXiv preprint arXiv:2311.10122}.

\bibitem[{Lin et~al.(2024)Lin, Zhang, Huang, Liu, Wen, and
  Peng}]{lin2024norton}
Yijie Lin, Jie Zhang, Zhenyu Huang, Jia Liu, Zujie Wen, and Xi~Peng. 2024.
\newblock Multi-granularity correspondence learning from long-term noisy
  videos.
\newblock In \emph{Proceedings of the International Conference on Learning
  Representations}.

\bibitem[{Liu et~al.(2024{\natexlab{a}})Liu, Li, Li, and Lee}]{Liu_2024_CVPR}
Haotian Liu, Chunyuan Li, Yuheng Li, and Yong~Jae Lee. 2024{\natexlab{a}}.
\newblock Improved baselines with visual instruction tuning.
\newblock In \emph{Proceedings of the IEEE/CVF Conference on Computer Vision
  and Pattern Recognition (CVPR)}, pages 26296--26306.

\bibitem[{Liu et~al.(2024{\natexlab{b}})Liu, Li, Li, Li, Zhang, Shen, and
  Lee}]{liu2024llavanext}
Haotian Liu, Chunyuan Li, Yuheng Li, Bo~Li, Yuanhan Zhang, Sheng Shen, and
  Yong~Jae Lee. 2024{\natexlab{b}}.
\newblock \href {https://llava-vl.github.io/blog/2024-01-30-llava-next/}
  {Llava-next: Improved reasoning, ocr, and world knowledge}.

\bibitem[{Liu et~al.(2024{\natexlab{c}})Liu, Li, Ge, Li, Shan, and
  Li}]{10658172}
Ruyang Liu, Chen Li, Yixiao Ge, Thomas~H. Li, Ying Shan, and Ge~Li.
  2024{\natexlab{c}}.
\newblock \href {https://doi.org/10.1109/CVPR52733.2024.01296} {Bt-adapter:
  Video conversation is feasible without video instruction tuning}.
\newblock In \emph{2024 IEEE/CVF Conference on Computer Vision and Pattern
  Recognition (CVPR)}, pages 13658--13667.

\bibitem[{Ma et~al.(2024)Ma, Jin, Wang, Xian, Feng, and Yang}]{Ma_2024_CVPR}
Fan Ma, Xiaojie Jin, Heng Wang, Yuchen Xian, Jiashi Feng, and Yi~Yang. 2024.
\newblock Vista-llama: Reducing hallucination in video language models via
  equal distance to visual tokens.
\newblock In \emph{Proceedings of the IEEE/CVF Conference on Computer Vision
  and Pattern Recognition (CVPR)}, pages 13151--13160.

\bibitem[{Maaz et~al.(2024)Maaz, Rasheed, Khan, and
  Khan}]{maaz-etal-2024-video}
Muhammad Maaz, Hanoona Rasheed, Salman Khan, and Fahad Khan. 2024.
\newblock \href {https://aclanthology.org/2024.acl-long.679}
  {Video-{C}hat{GPT}: Towards detailed video understanding via large vision and
  language models}.
\newblock In \emph{Proceedings of the 62nd Annual Meeting of the Association
  for Computational Linguistics (Volume 1: Long Papers)}, pages 12585--12602,
  Bangkok, Thailand. Association for Computational Linguistics.

\bibitem[{Mangalam et~al.(2023)Mangalam, Akshulakov, and Malik}]{egoschema}
Karttikeya Mangalam, Raiymbek Akshulakov, and Jitendra Malik. 2023.
\newblock \href {https://arxiv.org/abs/2308.09126} {Egoschema: A diagnostic
  benchmark for very long-form video language understanding}.
\newblock \emph{Preprint}, arXiv:2308.09126.

\bibitem[{Mayer and
  Moreno(2003)}]{NineWaysToReduceCognitiveLoadInMultimediaLearningNO4}
Richard~E. Mayer and Roxana Moreno. 2003.
\newblock Nine ways to reduce cognitive load in multimedia learning.
\newblock \emph{Educational Psychologist}, 38(1):43--52.

\bibitem[{Newell and Simon(1972)}]{HumanProblemSolvingNO12}
Allen Newell and Herbert~A. Simon. 1972.
\newblock \emph{Human Problem Solving}.
\newblock Prentice-Hall, Englewood Cliffs, NJ.

\bibitem[{OpenAI et~al.(2024{\natexlab{a}})OpenAI, :, Hurst, Lerer, Goucher,
  Perelman, Ramesh, Clark, Ostrow, Welihinda, Hayes, Radford, Mądry,
  Baker-Whitcomb, Beutel, Borzunov, Carney, Chow, Kirillov, Nichol, Paino,
  Renzin, Passos, Kirillov, Christakis, Conneau, Kamali, Jabri, Moyer, Tam,
  Crookes, Tootoochian, Tootoonchian, Kumar, Vallone, Karpathy, Braunstein,
  Cann, Codispoti, Galu, Kondrich, Tulloch, Mishchenko, Baek, Jiang, Pelisse,
  Woodford, Gosalia, Dhar, Pantuliano, Nayak, Oliver, Zoph, Ghorbani,
  Leimberger, Rossen, Sokolowsky, Wang, Zweig, Hoover, Samic, McGrew, Spero,
  Giertler, Cheng, Lightcap, Walkin, Quinn, Guarraci, Hsu, Kellogg, Eastman,
  Lugaresi, Wainwright, Bassin, Hudson, Chu, Nelson, Li, Shern, Conger,
  Barette, Voss, Ding, Lu, Zhang, Beaumont, Hallacy, Koch, Gibson, Kim, Choi,
  McLeavey, Hesse, Fischer, Winter, Czarnecki, Jarvis, Wei, Koumouzelis,
  Sherburn, Kappler, Levin, Levy, Carr, Farhi, Mely, Robinson, Sasaki, Jin,
  Valladares, Tsipras, Li, Nguyen, Findlay, Oiwoh, Wong, Asdar, Proehl, Yang,
  Antonow, Kramer, Peterson, Sigler, Wallace, Brevdo, Mays, Khorasani, Such,
  Raso, Zhang, von Lohmann, Sulit, Goh, Oden, Salmon, Starace, Brockman,
  Salman, Bao, Hu, Wong, Wang, Schmidt, Whitney, Jun, Kirchner,
  de~Oliveira~Pinto, Ren, Chang, Chung, Kivlichan, O'Connell, O'Connell,
  Osband, Silber, Sohl, Okuyucu, Lan, Kostrikov, Sutskever, Kanitscheider,
  Gulrajani, Coxon, Menick, Pachocki, Aung, Betker, Crooks, Lennon, Kiros,
  Leike, Park, Kwon, Phang, Teplitz, Wei, Wolfe, Chen, Harris, Varavva, Lee,
  Shieh, Lin, Yu, Weng, Tang, Yu, Jang, Candela, Beutler, Landers, Parish,
  Heidecke, Schulman, Lachman, McKay, Uesato, Ward, Kim, Huizinga, Sitkin,
  Kraaijeveld, Gross, Kaplan, Snyder, Achiam, Jiao, Lee, Zhuang, Harriman,
  Fricke, Hayashi, Singhal, Shi, Karthik, Wood, Rimbach, Hsu, Nguyen,
  Gu-Lemberg, Button, Liu, Howe, Muthukumar, Luther, Ahmad, Kai, Itow, Workman,
  Pathak, Chen, Jing, Guy, Fedus, Zhou, Mamitsuka, Weng, McCallum, Held,
  Ouyang, Feuvrier, Zhang, Kondraciuk, Kaiser, Hewitt, Metz, Doshi, Aflak,
  Simens, Boyd, Thompson, Dukhan, Chen, Gray, Hudnall, Zhang, Aljubeh, Litwin,
  Zeng, Johnson, Shetty, Gupta, Shah, Yatbaz, Yang, Zhong, Glaese, Chen,
  Janner, Lampe, Petrov, Wu, Wang, Fradin, Pokrass, Castro, de~Castro, Pavlov,
  Brundage, Wang, Khan, Murati, Bavarian, Lin, Yesildal, Soto, Gimelshein,
  Cone, Staudacher, Summers, LaFontaine, Chowdhury, Ryder, Stathas, Turley,
  Tezak, Felix, Kudige, Keskar, Deutsch, Bundick, Puckett, Nachum, Okelola,
  Boiko, Murk, Jaffe, Watkins, Godement, Campbell-Moore, Chao, McMillan, Belov,
  Su, Bak, Bakkum, Deng, Dolan, Hoeschele, Welinder, Tillet, Pronin, Tillet,
  Dhariwal, Yuan, Dias, Lim, Arora, Troll, Lin, Lopes, Puri, Miyara, Leike,
  Gaubert, Zamani, Wang, Donnelly, Honsby, Smith, Sahai, Ramchandani, Huet,
  Carmichael, Zellers, Chen, Chen, Nigmatullin, Cheu, Jain, Altman, Schoenholz,
  Toizer, Miserendino, Agarwal, Culver, Ethersmith, Gray, Grove, Metzger,
  Hermani, Jain, Zhao, Wu, Jomoto, Wu, Shuaiqi, Xia, Phene, Papay, Narayanan,
  Coffey, Lee, Hall, Balaji, Broda, Stramer, Xu, Gogineni, Christianson,
  Sanders, Patwardhan, Cunninghman, Degry, Dimson, Raoux, Shadwell, Zheng,
  Underwood, Markov, Sherbakov, Rubin, Stasi, Kaftan, Heywood, Peterson,
  Walters, Eloundou, Qi, Moeller, Monaco, Kuo, Fomenko, Chang, Zheng, Zhou,
  Manassra, Sheu, Zaremba, Patil, Qian, Kim, Cheng, Zhang, He, Zhang, Jin, Dai,
  and Malkov}]{gpt4o}
OpenAI, :, Aaron Hurst, Adam Lerer, Adam~P. Goucher, Adam Perelman, Aditya
  Ramesh, Aidan Clark, AJ~Ostrow, Akila Welihinda, Alan Hayes, Alec Radford,
  Aleksander Mądry, Alex Baker-Whitcomb, Alex Beutel, Alex Borzunov, Alex
  Carney, Alex Chow, Alex Kirillov, and 401 others. 2024{\natexlab{a}}.
\newblock \href {https://arxiv.org/abs/2410.21276} {Gpt-4o system card}.
\newblock \emph{Preprint}, arXiv:2410.21276.

\bibitem[{OpenAI et~al.(2024{\natexlab{b}})OpenAI, Achiam, Adler, Agarwal,
  Ahmad, Akkaya, Aleman, Almeida, Altenschmidt, Altman, Anadkat, Avila,
  Babuschkin, Balaji, Balcom, Baltescu, Bao, Bavarian, Belgum, Bello, Berdine,
  Bernadett-Shapiro, Berner, Bogdonoff, Boiko, Boyd, Brakman, Brockman, Brooks,
  Brundage, Button, Cai, Campbell, Cann, Carey, Carlson, Carmichael, Chan,
  Chang, Chantzis, Chen, Chen, Chen, Chen, Chen, Chess, Cho, Chu, Chung,
  Cummings, Currier, Dai, Decareaux, Degry, Deutsch, Deville, Dhar, Dohan,
  Dowling, Dunning, Ecoffet, Eleti, Eloundou, Farhi, Fedus, Felix, Fishman,
  Forte, Fulford, Gao, Georges, Gibson, Goel, Gogineni, Goh, Gontijo-Lopes,
  Gordon, Grafstein, Gray, Greene, Gross, Gu, Guo, Hallacy, Han, Harris, He,
  Heaton, Heidecke, Hesse, Hickey, Hickey, Hoeschele, Houghton, Hsu, Hu, Hu,
  Huizinga, Jain, Jain, Jang, Jiang, Jiang, Jin, Jin, Jomoto, Jonn, Jun,
  Kaftan, Łukasz Kaiser, Kamali, Kanitscheider, Keskar, Khan, Kilpatrick, Kim,
  Kim, Kim, Kirchner, Kiros, Knight, Kokotajlo, Łukasz Kondraciuk, Kondrich,
  Konstantinidis, Kosic, Krueger, Kuo, Lampe, Lan, Lee, Leike, Leung, Levy, Li,
  Lim, Lin, Lin, Litwin, Lopez, Lowe, Lue, Makanju, Malfacini, Manning, Markov,
  Markovski, Martin, Mayer, Mayne, McGrew, McKinney, McLeavey, McMillan,
  McNeil, Medina, Mehta, Menick, Metz, Mishchenko, Mishkin, Monaco, Morikawa,
  Mossing, Mu, Murati, Murk, Mély, Nair, Nakano, Nayak, Neelakantan, Ngo, Noh,
  Ouyang, O'Keefe, Pachocki, Paino, Palermo, Pantuliano, Parascandolo, Parish,
  Parparita, Passos, Pavlov, Peng, Perelman, de~Avila Belbute~Peres, Petrov,
  de~Oliveira~Pinto, Michael, Pokorny, Pokrass, Pong, Powell, Power, Power,
  Proehl, Puri, Radford, Rae, Ramesh, Raymond, Real, Rimbach, Ross, Rotsted,
  Roussez, Ryder, Saltarelli, Sanders, Santurkar, Sastry, Schmidt, Schnurr,
  Schulman, Selsam, Sheppard, Sherbakov, Shieh, Shoker, Shyam, Sidor, Sigler,
  Simens, Sitkin, Slama, Sohl, Sokolowsky, Song, Staudacher, Such, Summers,
  Sutskever, Tang, Tezak, Thompson, Tillet, Tootoonchian, Tseng, Tuggle,
  Turley, Tworek, Uribe, Vallone, Vijayvergiya, Voss, Wainwright, Wang, Wang,
  Wang, Ward, Wei, Weinmann, Welihinda, Welinder, Weng, Weng, Wiethoff,
  Willner, Winter, Wolrich, Wong, Workman, Wu, Wu, Wu, Xiao, Xu, Yoo, Yu, Yuan,
  Zaremba, Zellers, Zhang, Zhang, Zhao, Zheng, Zhuang, Zhuk, and Zoph}]{gpt4}
OpenAI, Josh Achiam, Steven Adler, Sandhini Agarwal, Lama Ahmad, Ilge Akkaya,
  Florencia~Leoni Aleman, Diogo Almeida, Janko Altenschmidt, Sam Altman,
  Shyamal Anadkat, Red Avila, Igor Babuschkin, Suchir Balaji, Valerie Balcom,
  Paul Baltescu, Haiming Bao, Mohammad Bavarian, Jeff Belgum, and 262 others.
  2024{\natexlab{b}}.
\newblock \href {https://arxiv.org/abs/2303.08774} {Gpt-4 technical report}.
\newblock \emph{Preprint}, arXiv:2303.08774.

\bibitem[{Paas et~al.(2003)Paas, Renkl, and
  Sweller}]{ReducingExtraneousLoadImportantNO15}
Fred Paas, Alexander Renkl, and John Sweller. 2003.
\newblock \href {https://doi.org/10.1207/S15326985EP3801\_1} {Cognitive load
  theory and instructional design: Recent developments}.
\newblock \emph{Educational Psychologist}, 38(1):1--4.

\bibitem[{Radford et~al.(2021)Radford, Kim, Hallacy, Ramesh, Goh, Agarwal,
  Sastry, Askell, Mishkin, Clark, Krueger, and Sutskever}]{clip}
Alec Radford, Jong~Wook Kim, Chris Hallacy, Aditya Ramesh, Gabriel Goh,
  Sandhini Agarwal, Girish Sastry, Amanda Askell, Pamela Mishkin, Jack Clark,
  Gretchen Krueger, and Ilya Sutskever. 2021.
\newblock \href {https://arxiv.org/abs/2103.00020} {Learning transferable
  visual models from natural language supervision}.
\newblock \emph{Preprint}, arXiv:2103.00020.

\bibitem[{Shang et~al.(2024)Shang, You, Subramanian, Darrell, and
  Herzig}]{shang-etal-2024-traveler}
Chuyi Shang, Amos You, Sanjay Subramanian, Trevor Darrell, and Roei Herzig.
  2024.
\newblock \href {https://doi.org/10.18653/v1/2024.emnlp-main.544}
  {{T}rave{LER}: A modular multi-{LMM} agent framework for video
  question-answering}.
\newblock In \emph{Proceedings of the 2024 Conference on Empirical Methods in
  Natural Language Processing}, pages 9740--9766, Miami, Florida, USA.
  Association for Computational Linguistics.

\bibitem[{Sweller(1988)}]{cognitiveLoadDuringProblemSolving:EffectsOnLearningNO3}
John Sweller. 1988.
\newblock Cognitive load during problem solving: Effects on learning.
\newblock \emph{Cognitive Science}, 12(2):257--285.

\bibitem[{Tishby et~al.(1999)Tishby, Pereira, and
  Bialek}]{tishby1999information}
Naftali Tishby, Fernando~C. Pereira, and William Bialek. 1999.
\newblock The information bottleneck method.
\newblock In \emph{Proceedings of the 37th Annual Allerton Conference on
  Communication, Control, and Computing}, pages 368--377.

\bibitem[{Tishby and Zaslavsky(2015)}]{tishby2015deep}
Naftali Tishby and Noga Zaslavsky. 2015.
\newblock Deep learning and the information bottleneck principle.
\newblock In \emph{IEEE Information Theory Workshop (ITW)}, pages 1--5.

\bibitem[{Treisman and Gelade(1980)}]{FITNO1}
Anne~M. Treisman and Garry Gelade. 1980.
\newblock \href {https://doi.org/10.1016/0010-0285(80)90005-5} {A
  feature-integration theory of attention}.
\newblock \emph{Cognitive Psychology}, 12(1):97--136.

\bibitem[{Wang et~al.(2024{\natexlab{a}})Wang, He, Hong, Cheng, Zhang, Qi,
  Huang, Xu, Dong, Ding, and Tang}]{lvbench}
Weihan Wang, Zehai He, Wenyi Hong, Yean Cheng, Xiaohan Zhang, Ji~Qi, Shiyu
  Huang, Bin Xu, Yuxiao Dong, Ming Ding, and Jie Tang. 2024{\natexlab{a}}.
\newblock \href {https://arxiv.org/abs/2406.08035} {Lvbench: An extreme long
  video understanding benchmark}.
\newblock \emph{Preprint}, arXiv:2406.08035.

\bibitem[{Wang et~al.(2025)Wang, Gao, Gu, Pu, Cui, Wei, Liu, Jing, Ye, Shao,
  Wang, Chen, Zhang, Yang, Wang, Wei, Yin, Li, Cui, Chen, Ding, Tian, Wu, Xie,
  Li, Yang, Duan, Wang, Hou, Hao, Zhang, Li, Zhao, Duan, Deng, Fu, He, Wang,
  He, Shi, He, Xiong, Lv, Wu, Shao, Zhang, Deng, Qi, Ge, Guo, Zhang, Zhang,
  Cao, Lin, Tang, Gao, Huang, Gu, Lyu, Tang, Wang, Lv, Ouyang, Wang, Dou, Zhu,
  Lu, Lin, Dai, Su, Zhou, Chen, Qiao, Wang, and Luo}]{InternVL3}
Weiyun Wang, Zhangwei Gao, Lixin Gu, Hengjun Pu, Long Cui, Xingguang Wei,
  Zhaoyang Liu, Linglin Jing, Shenglong Ye, Jie Shao, Zhaokai Wang, Zhe Chen,
  Hongjie Zhang, Ganlin Yang, Haomin Wang, Qi~Wei, Jinhui Yin, Wenhao Li, Erfei
  Cui, and 56 others. 2025.
\newblock \href {https://arxiv.org/abs/2508.18265} {Internvl3.5: Advancing
  open-source multimodal models in versatility, reasoning, and efficiency}.
\newblock \emph{Preprint}, arXiv:2508.18265.

\bibitem[{Wang et~al.(2024{\natexlab{b}})Wang, Zhang, Zohar, and
  Yeung-Levy}]{VideoAgent}
Xiaohan Wang, Yuhui Zhang, Orr Zohar, and Serena Yeung-Levy.
  2024{\natexlab{b}}.
\newblock Videoagent: Long-form video understanding with large language model
  as agent.
\newblock \emph{European Conference on Computer Vision (ECCV)}.

\bibitem[{Wang et~al.(2024{\natexlab{c}})Wang, Yu, Stengel-Eskin, Yoon, Cheng,
  Bertasius, and Bansal}]{wang2024videotreeadaptivetreebasedvideo}
Ziyang Wang, Shoubin Yu, Elias Stengel-Eskin, Jaehong Yoon, Feng Cheng, Gedas
  Bertasius, and Mohit Bansal. 2024{\natexlab{c}}.
\newblock \href {https://arxiv.org/abs/2405.19209} {Videotree: Adaptive
  tree-based video representation for llm reasoning on long videos}.
\newblock \emph{Preprint}, arXiv:2405.19209.

\bibitem[{Wei et~al.(2022)Wei, Wang, Schuurmans, Bosma, Ichter, Xia, Chi, Le,
  and Zhou}]{wei2022chain}
Jason Wei, Xuezhi Wang, Dale Schuurmans, Maarten Bosma, Brian Ichter, Fei Xia,
  Ed~H. Chi, Quoc~V. Le, and Denny Zhou. 2022.
\newblock Chain-of-thought prompting elicits reasoning in large language
  models.
\newblock In \emph{Advances in Neural Information Processing Systems
  (NeurIPS)}.

\bibitem[{Wolfe(1994)}]{GS2.0NO2}
Jeremy~M. Wolfe. 1994.
\newblock Guided search 2.0: A revised model of visual search.
\newblock \emph{Psychonomic Bulletin \& Review}, 1(2):202--238.

\bibitem[{Wu and Xie(2024)}]{wu2024v}
Penghao Wu and Saining Xie. 2024.
\newblock V?: Guided visual search as a core mechanism in multimodal llms.
\newblock In \emph{Proceedings of the IEEE/CVF Conference on Computer Vision
  and Pattern Recognition}, pages 13084--13094.

\bibitem[{Xiao et~al.(2021)Xiao, Shang, Yao, and Chua}]{xiao2021next}
Junbin Xiao, Xindi Shang, Angela Yao, and Tat-Seng Chua. 2021.
\newblock Next-qa: Next phase of question-answering to explaining temporal
  actions.
\newblock In \emph{Proceedings of the IEEE/CVF Conference on Computer Vision
  and Pattern Recognition (CVPR)}, pages 9777--9786.

\bibitem[{Xiao et~al.(2024)Xiao, Yao, Li, and Chua}]{nextgqa}
Junbin Xiao, Angela Yao, Yicong Li, and Tat-Seng Chua. 2024.
\newblock Can i trust your answer? visually grounded video question answering.
\newblock In \emph{Proceedings of the IEEE/CVF Conference on Computer Vision
  and Pattern Recognition}, pages 13204--13214.

\bibitem[{Xu et~al.(2019)Xu, He, Plummer, Sigal, Sclaroff, and
  Saenko}]{xu2019multilevel}
Huijuan Xu, Kun He, Bryan~A. Plummer, Leonid Sigal, Stan Sclaroff, and Kate
  Saenko. 2019.
\newblock Multilevel language and vision integration for text-to-clip
  retrieval.
\newblock In \emph{AAAI}.

\bibitem[{Xu et~al.(2024{\natexlab{a}})Xu, Zhao, Zhou, Lin, Ng, and
  Feng}]{Xu2024PLLaVAP}
Lin Xu, Yilin Zhao, Daquan Zhou, Zhijie Lin, See~Kiong Ng, and Jiashi Feng.
  2024{\natexlab{a}}.
\newblock \href {https://api.semanticscholar.org/CorpusID:269430328} {Pllava :
  Parameter-free llava extension from images to videos for video dense
  captioning}.
\newblock \emph{ArXiv}, abs/2404.16994.

\bibitem[{Xu et~al.(2024{\natexlab{b}})Xu, Sun, Xie, Zhai, and Du}]{vtggpt}
Yifang Xu, Yunzhuo Sun, Zien Xie, Benxiang Zhai, and Sidan Du.
  2024{\natexlab{b}}.
\newblock Vtg-gpt: Tuning-free zero-shot video temporal grounding with gpt.
\newblock \emph{Applied Sciences}, 14(5):1894.

\bibitem[{Yao et~al.(2023)Yao, Yu, Zhao, Shafran, Griffiths, Cao, and
  Narasimhan}]{yao2023tree}
Shunyu Yao, Dian Yu, Jeffrey Zhao, Izhak Shafran, Thomas~L. Griffiths, Yuan
  Cao, and Karthik Narasimhan. 2023.
\newblock Tree of thoughts: Deliberate problem solving with large language
  models.
\newblock In \emph{Advances in Neural Information Processing Systems
  (NeurIPS)}.

\bibitem[{Yu et~al.(2019)Yu, Xu, Yu, Yu, Zhao, Zhuang, and
  Tao}]{yu2019activityqa}
Zhou Yu, Dejing Xu, Jun Yu, Ting Yu, Zhou Zhao, Yueting Zhuang, and Dacheng
  Tao. 2019.
\newblock Activitynet-qa: A dataset for understanding complex web videos via
  question answering.
\newblock In \emph{AAAI}, pages 9127--9134.

\bibitem[{Yuan et~al.(2019)Yuan, Mei, and Zhu}]{10.1609/aaai.v33i01.33019159}
Yitian Yuan, Tao Mei, and Wenwu Zhu. 2019.
\newblock \href {https://doi.org/10.1609/aaai.v33i01.33019159} {To find where
  you talk: temporal sentence localization in video with attention based
  location regression}.
\newblock In \emph{Proceedings of the Thirty-Third AAAI Conference on
  Artificial Intelligence and Thirty-First Innovative Applications of
  Artificial Intelligence Conference and Ninth AAAI Symposium on Educational
  Advances in Artificial Intelligence}, AAAI'19/IAAI'19/EAAI'19. AAAI Press.

\bibitem[{Zacks et~al.(2007)Zacks, Speer, Swallow, Braver, and
  Reynolds}]{zacks2007event}
Jeffrey~M. Zacks, Nicole~K. Speer, Khena~M. Swallow, Todd~S. Braver, and
  Jeremy~R. Reynolds. 2007.
\newblock \href {https://doi.org/10.1037/0033-2909.133.2.273} {Event
  perception: A mind/brain perspective}.
\newblock \emph{Psychological Bulletin}, 133(2):273--293.

\bibitem[{Zhang et~al.(2020)Zhang, Sun, Jing, and Zhou}]{zhang-etal-2020-span}
Hao Zhang, Aixin Sun, Wei Jing, and Joey~Tianyi Zhou. 2020.
\newblock \href {https://doi.org/10.18653/v1/2020.acl-main.585} {Span-based
  localizing network for natural language video localization}.
\newblock In \emph{Proceedings of the 58th Annual Meeting of the Association
  for Computational Linguistics}, pages 6543--6554, Online. Association for
  Computational Linguistics.

\bibitem[{Zhang et~al.(2025)Zhang, Li, Qian, Liu, Yu, Han, Fung, McKeown, Zhai,
  Li et~al.}]{knowledgeovershadowing}
Yuji Zhang, Sha Li, Cheng Qian, Jiateng Liu, Pengfei Yu, Chi Han, Yi~R Fung,
  Kathleen McKeown, Chengxiang Zhai, Manling Li, and 1 others. 2025.
\newblock The law of knowledge overshadowing: Towards understanding,
  predicting, and preventing llm hallucination.
\newblock \emph{arXiv preprint arXiv:2502.16143}.

\bibitem[{Zheng et~al.(2024)Zheng, Cai, Chen, Peng, and Liu}]{tfvtg}
Minghang Zheng, Xinhao Cai, Qingchao Chen, Yuxin Peng, and Yang Liu. 2024.
\newblock \href {https://arxiv.org/abs/2408.16219} {Training-free video
  temporal grounding using large-scale pre-trained models}.
\newblock \emph{Preprint}, arXiv:2408.16219.

\bibitem[{Zhou et~al.(2024{\natexlab{a}})Zhou, Shu, Zhao, Wu, Xiao, Yang,
  Xiong, Zhang, Huang, and Liu}]{mlvu}
Junjie Zhou, Yan Shu, Bo~Zhao, Boya Wu, Shitao Xiao, Xi~Yang, Yongping Xiong,
  Bo~Zhang, Tiejun Huang, and Zheng Liu. 2024{\natexlab{a}}.
\newblock Mlvu: A comprehensive benchmark for multi-task long video
  understanding.
\newblock \emph{arXiv preprint arXiv:2406.04264}.

\bibitem[{Zhou et~al.(2018)Zhou, Xu, and Corso}]{ZhXuCoAAAI18}
Luowei Zhou, Chenliang Xu, and Jason~J Corso. 2018.
\newblock \href
  {https://www.aaai.org/ocs/index.php/AAAI/AAAI18/paper/view/17344} {Towards
  automatic learning of procedures from web instructional videos}.
\newblock In \emph{AAAI Conference on Artificial Intelligence}, pages
  7590--7598.

\bibitem[{Zhou et~al.(2024{\natexlab{b}})Zhou, Arnab, Buch, Yan, Myers, Xiong,
  Nagrani, and Schmid}]{clustering_knn}
Xingyi Zhou, Anurag Arnab, Shyamal Buch, Shen Yan, Austin Myers, Xuehan Xiong,
  Arsha Nagrani, and Cordelia Schmid. 2024{\natexlab{b}}.
\newblock Streaming dense video captioning.
\newblock In \emph{Proceedings of the IEEE/CVF Conference on Computer Vision
  and Pattern Recognition (CVPR)}, pages 18243--18252.

\end{thebibliography}

\appendix

\label{sec:appendix}

\begin{table*}[ht]
  \centering

  \begin{subtable}{\textwidth}
    \centering
    \resizebox{\textwidth}{!}{%
    \begin{tabular}{c|cccc|cccc|cccc|cccc}
      \toprule
      \textbf{\multirow{2}{*}{Method}}  &\multicolumn{4}{c}{Temporal Relational }  & \multicolumn{4}{c}{Timepoint Indexed} & \multicolumn{4}{c}{Multifaceted Integrative}  & \multicolumn{4}{c}{\textbf{Average}}\\
       & mIoU & Pre. & Cov. & F1 & mIoU & Pre. & Cov. & F1 & mIoU & Pre. & Cov. & F1 & mIoU & Pre. & Cov. & F1 \\
       \midrule
       VTG-GPT \citep{vtggpt} & 0.17 & 0.26 & 0.32 & 0.29 & 0.15 & 0.22 & 0.30 & 0.26 & 0.16 & 0.23 & 0.31 & 0.24 & 0.16 & 0.27 & 0.31 & 0.27 \\
       \citet{tfvtg} & 0.11 & 0.18 & 0.21 & 0.19 & 0.04 & 0.08 & 0.09 & 0.08 & 0.06 & 0.10 & 0.11 & 0.10 & 0.07 & 0.12 & 0.13 & 0.12 \\
       	\textbf{ReSimplifyIt} (Ours) & \textbf{0.23} & \textbf{0.36} & \textbf{0.39} & \textbf{0.37} & \textbf{0.98} & \textbf{1.0} & \textbf{0.98} & \textbf{0.99} & \textbf{0.47} & \textbf{0.60} & \textbf{0.69} & \textbf{0.64} & \textbf{0.56} & \textbf{0.65} & \textbf{0.69} & \textbf{0.67}\\
      \bottomrule
    \end{tabular}
    }
    \caption{more results on stage 1 video output}
    \label{tab:emc_main_more_vid}
  \end{subtable}

  \caption{More results on Stage-1 evaluation on EMCompress dataset. 'Pre.' and 'Cov.' stands for 'precision' and 'coverage'.}
  \label{tab:stage1_eval_appendix}
\end{table*}

\begin{figure*}[t]
  \includegraphics[width=\textwidth]{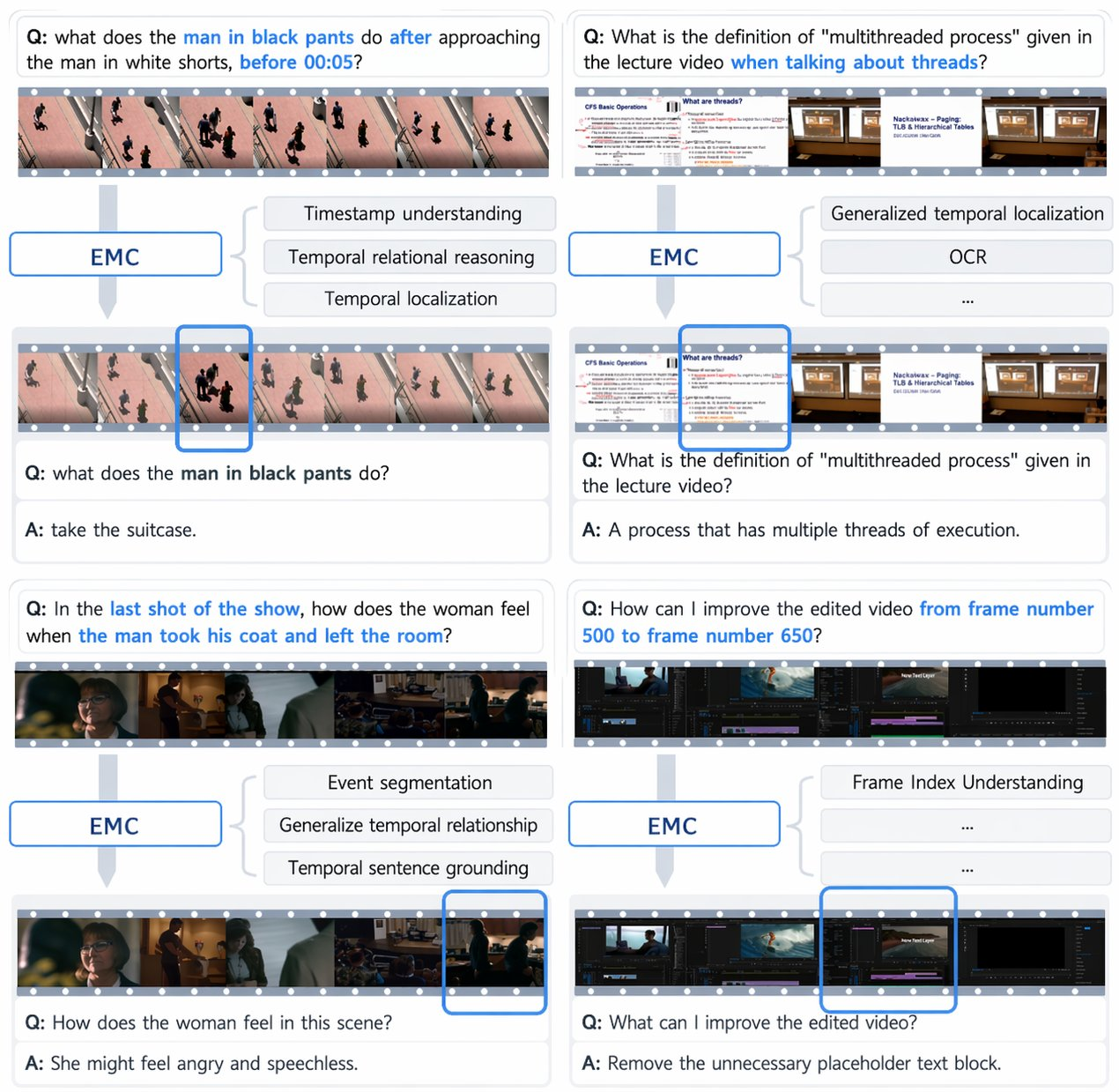}
  \caption{More examples of the EMC task.\vspace{-10pt}}
  \label{fig:emc_task_figure}
\end{figure*}

\begin{figure*}[htbp] 
    \centering
    \begin{minipage}{\textwidth} 
\begin{algorithm}[H] 
    \caption{\emph{ISODATA Clustering}}
    \begin{algorithmic}[1]
        \Require Data matrix $\mathbf{X} \in \mathbb{R}^{n \times d}$, Candidate frames $\mathbf{F} \in \mathbb{R}^{n \times c \times w \times h}$, frame feature extractor $Enc()$, initial cluster count $k$, max iterations $T$, minimum intra-cluster similarity $\theta_{split}$, maximum inter-cluster similarity $\theta_{merge}$, max clusters $k_{max}$, min clusters $k_{min}$, max center shift $\delta_{max}$, min elements per cluster $n_{min}$
        \Ensure Cluster assignments $\mathbf{C}$, Final cluster centers $\mathbf{M}$

        \State $\mathbf{X} \gets \text{Enc}(\mathbf{F})$
        \State $\mathbf{X}_{norm} \gets \text{normalize}(\mathbf{X}, \ell_2)$
        \State $\mathbf{M} \gets \text{random\_sample}(\mathbf{X}, k)$
        \State $t \gets 0$
        \Repeat
            \State Compute cosine similarity matrix $\mathbf{S} = \mathbf{X}_{norm} \mathbf{M}_{norm}^T$
            \State $\mathbf{C} \gets \arg\max(\mathbf{S}, \text{axis}=1)$ \Comment{Assign points to nearest clusters}
            
            \For{each cluster $i$}
                \State Update center: $\mathbf{m}_i \gets \text{mean}(\mathbf{X}[\mathbf{C}==i])$
            \EndFor
            
            \If{any cluster size $< n_{min}$}
                \State Merge smallest cluster with nearest neighbor \Comment{Minimum elements enforcement}
            \EndIf
            
            \For{each cluster $i$}
                \If{intra-cluster similarity$(\mathbf{X}[\mathbf{C}==i], \mathbf{m}_i) < \theta_{split}$ \textbf{and} $k < k_{max}$}
                    \State Split cluster $i$ into two new clusters \Comment{Splitting phase}
                    \State $k \gets k + 1$
                \EndIf
            \EndFor
            
            \For{all cluster pairs $(i,j)$}
                \If{inter-cluster similarity$(\mathbf{m}_i, \mathbf{m}_j) > \theta_{merge}$ \textbf{and} $k > k_{min}$}
                    \State Merge clusters $i$ and $j$ \Comment{Merging phase}
                    \State $k \gets k - 1$
                \EndIf
            \EndFor
            
            \State Compute center shifts $\Delta \mathbf{M} = \|\mathbf{M}_{new} - \mathbf{M}_{old}\|$
            \State $t \gets t + 1$
        \Until{$t \geq T$ \textbf{or} $\max(\Delta \mathbf{M}) < \delta_{max}$}
        
        \State \Return $\mathbf{C}, \mathbf{M}$
    \end{algorithmic}
\label{alg_isodata}
\end{algorithm}
    \end{minipage}
\end{figure*}

\begin{figure*}[htbp] 
    
    \centering
    \begin{minipage}{\textwidth} 
        \begin{algorithm}[H] 
            \caption{\emph{ReSimplifyIt}}
            \label{alg_ReSimplify}
            \small
            \begin{algorithmic}[1]
                \Require Video $v$, question $q$
                \Ensure Compressed video $V'$, compressed question $q'$

                \State $V\_copy, q\_copy \gets V, q$
                \State $success\_history, failure\_history \gets \texttt{SuccessHistory()}, \texttt{FailureHistory()}$
                \While{true} 
                    \State $launcher, validator, viewer \gets \texttt{Launcher()}, \texttt{Validator()}, \texttt{Viewer()}$
                    \State $decision, q', trimming\_instruction \gets \texttt{launcher}(q\_copy, success\_history, failure\_history)$
                    \If{$decision=="proceed"$}
                        \State $judgement, request, result, reason \gets \texttt{validator}(q\_copy, q', trimming\_instruction)$
                        \While{$judgement=="view"$}
                            \State $response \gets \texttt{viewer}(V', request)$
                            \State $\texttt{validator.read\_response}(response)$
                        \EndWhile
                        \If{$judgement=="succeeded"$}
                            \State $V' \gets result$
                            \State $success\_history.append([q\_copy, q', trimming\_instruction])$
                            \State $failure\_history.empty()$
                            \State $V\_copy, q\_copy \gets V', q'$
                        \Else
                            \State $failure\_history.append([q\_copy, q', trimming\_instruction])$
                        \EndIf
                    \Else
                        \State \Return $V\_copy, q\_copy$
                    \EndIf
                \EndWhile
            \end{algorithmic}
        \label{alg_emc}
        \end{algorithm}
        
    \end{minipage}
\end{figure*}

\begin{table*}
  \centering
  \resizebox{0.7\linewidth}{!}{\begin{tabular}{c|cc}
    \toprule
    \textbf{\multirow{2}{*}{Method}} & \multicolumn{2}{c}{ActivityNetQA} \\ 
    & w/emc & w/o \\ 
    \midrule
    Video-ChatGPT \citep{maaz-etal-2024-video} & 58.5 & 50.5\\ 
    Video-LLaVA \citep{lin2023video} & 61.2 & 52.1 \\ 
    ChatUniVi \citep{chatunivi} & 63.2 & 52.0 \\
    LLaVA-NExT \citep{liu2024llavanext} & 67.8 & 62.5 \\ 
    VideoAgent \citep{VideoAgent} & 61.5 & 60.2 \\ 
    VideoTree \citep{wang2024videotreeadaptivetreebasedvideo} & 59.0 & 63.6 \\ 
    \bottomrule
  \end{tabular}}
  
  \resizebox{\linewidth}{!}{\begin{tabular}{c|cccccccc}
    \toprule
    \textbf{\multirow{3}{*}{Method}} & \multicolumn{8}{c}{EMCompress} \\ 
    & \multicolumn{2}{c}{TRR.} & \multicolumn{2}{c}{TIR.} & \multicolumn{2}{c}{MIR.} & \multicolumn{2}{c}{Avg.} \\ 
    & w/emc & w/o & w/emc & w/o & w/emc & w/o & w/emc & w/o \\
    \midrule 
    Video-ChatGPT \citep{maaz-etal-2024-video} & 31.7 & 30.5 & 45.2 & 26.5 & 39.8 & 29.73 & 38.9 & 28.91 \\ 
    Video-LLaVA \citep{lin2023video} & 39.3 & 37.4 & 46.8 & 28.2 & 43.5 & 31.6 & 43.2 & 32.4 \\ 
    ChatUniVi \citep{chatunivi} & 54.4 & 39.8 & 57.4 & 52.8 & 58.7 & 49.6 & 56.5 & 47.4 \\
    LLaVA-NExT \citep{liu2024llavanext} & 46.5 & 44.1 & 65.9 & 33.4 & 62.2 & 38.3 & 58.2 & 48.6 \\ 
    VideoAgent \citep{VideoAgent} & 30.2 & 48.7 & 46.1 & 55.2 & 38.9 & 57.8 & 38.4 & 53.9 \\ 
    VideoTree \citep{wang2024videotreeadaptivetreebasedvideo} & 59.3 & 64.0 & 59.2 & 72.2 & 52.8 & 72.0 & 57.1 & 69.4  \\ 
    \bottomrule
  \end{tabular}}
  
  \resizebox{\linewidth}{!}{\begin{tabular}{c|cccccccc}
    \toprule
    \textbf{\multirow{3}{*}{Method}} & \multicolumn{8}{c}{NExT-QA} \\ 
    & \multicolumn{2}{c}{Tem.} & \multicolumn{2}{c}{Cau.} & \multicolumn{2}{c}{Des.} & \multicolumn{2}{c}{Avg.} \\ 
    & w/emc & w/o & w/emc & w/o & w/emc & w/o & w/emc & w/o \\
    \midrule
    Video-ChatGPT \citep{maaz-etal-2024-video} & 45.5 & 23.7 & 45.5 & 56.7 & 45.6 & 43.0 & 45.5 & 44.2 \\
    Video-LLaVA \citep{lin2023video} & 42.8 & 42.8 & 50.4 & 48.9 & 55.1 & 44.5 & 52.2 & 46.3 \\ 
    ChatUniVi \citep{chatunivi} & - & - & - & - & - & - & 5 & 28 \\
    LLaVA-NExT \citep{liu2024llavanext} & 57.5 & 52.3 & 62.0 & 59.0 & 61.4 & 56.5 & 60.5 & 56.6 \\
    VideoAgent \citep{VideoAgent} & 49.2 & 47.3 & 43.6 & 41.9 & 51.7 & 51.1 & 47.0 & 45.0\\
    VideoTree \citep{wang2024videotreeadaptivetreebasedvideo} & 62.4 & 59.9 & 57.7 & 66.1 & 66.8 & 72.3 & 60.0 & 65.2 \\ 
    \bottomrule
  \end{tabular}}
  
  \resizebox{\linewidth}{!}{\begin{tabular}{c|cccccccc}
    \toprule
    \textbf{\multirow{3}{*}{Method}} & \multicolumn{8}{c}{NExT-OE} \\ 
    & \multicolumn{2}{c}{Tem.} & \multicolumn{2}{c}{Des.} & \multicolumn{2}{c}{Cau.} & \multicolumn{2}{c}{Avg.} \\ 
    & w/emc & w/o & w/emc & w/o & w/emc & w/o & w/emc & w/o \\ 
    \midrule
    Video-ChatGPT \citep{maaz-etal-2024-video} & 49.6 & 40.8 & 51.2 & 43.2 & 46.9 & 38.6 & 49.8 & 41.5 \\ 
    Video-LLaVA \citep{lin2023video} & 37.5 & 37.5 & 53.2 & 46.7 & 47.6 & 43.4 & 48.0 & 43.3 \\ 
    ChatUniVi \citep{chatunivi} & 30.8 & 30.8 & 36.9 & 23.1 & 34.0 & 26.4 & 34.5 & 25.9 \\ 
    LLaVA-NExT \citep{liu2024llavanext} & 41.4 & 39.11 & 59.7 & 46.0 & 50.1 & 44.6 & 52.3 & 43.7 \\ 
    VideoAgent \citep{VideoAgent} & 44.1 & 53.4 & 48.8 & 46.0 & 50.4 & 52.6 & 47.8 & 49.6\\ 
    VideoTree \citep{wang2024videotreeadaptivetreebasedvideo} & 52.1 & 61.1 & 58.0 & 60.7 & 64.4 & 65.6 & 57.7 & 61.9 \\ 
    \bottomrule
  \end{tabular}}
  \caption{Evaluation results of EMC-guided inference on four benchmark: ActivityNetQA, EMCompress, NExT-QA, and NExT-OE, with the sub-categories presented in each benchmark. }
  \label{stage2_eval_appendix}
\end{table*}

\section{Viewer Implementation Details}
\label{appendix_viewer}

\paragraph{Keyframe Extraction and Captioning.} 
We utilize a two-stage keyframe extraction process combined with frame captioning. In the first stage, we implement the MPEG-4 compression technique \citep{mpeg} to extract all I-frames as candidate keyframes. I-frames typically contain rich visual content and clarity, or represent scene transitions.

In the second stage, we apply a modified Isodata clustering algorithm to the visual features of these I-frames, selecting cluster centers as final keyframes. This algorithm adaptively determines the number of clusters, unlike KNN-based clustering \citep{wang2024videotreeadaptivetreebasedvideo, clustering_knn}, which requires a pre-defined number of clusters or external supervision. Our approach ensures generalizability and robustness for diverse video types.

After clustering, we utilize an off-the-shelf vision-language model (VLM) to generate frame captions for the selected keyframes. The overall process is denoted by:
\begin{center}
$C = prep(start, end)$
\end{center}
where $C$ is the set of frame captions between the given timestamps.

\paragraph{Scanning.} 
Given a timestamp range $(start, end)$, the Viewer calls an external LLM with $C$ to produce an overview caption. It may query additional frames via a captioning tool to refine its summary, mimicking how users drag to specific timestamps for clarification. The process is:
\begin{center}
$Cap = Scanner(prep(start, end), t)$
\end{center}
where $t$ is the task prompt.

\paragraph{Localizing.} 
To identify the timestamp range matching a textual description, we propose a lightweight three-stage search process:

\textbf{Stage 1:}  We feed an external LLM the set of keyframe captions to let it acquire an overall capture of the video content. At the same time, we instruct the LLM to output five most possible timestamps that depicts the text query. The LLM is able to call frame caption tool to acquire extra captions at arbitrary timestamps, before it's confident enough to output the answer. Formally:
\begin{center}
$P = stage_1(prep(0, d), e, t)$
\end{center}
where $e$ are extra captions and $t$ is the prompt. $P$ gives the resulting list of five timestamps which depicts the language query the best.

\textbf{Stage 2:} We initialize the conversation and feed it the frame captions at these five timestamps, while instructing it to pick the one timestamp that best depicts the language query, out of the five candidates. Also, the LLM is able to call frame caption tool to acquire extra captions at arbitrary timestamps before it's confident enough to output the answer. 
\begin{center}
$t_{best} = stage_2(P, e, t)$
\end{center}
where $P$ is the output of the previous step, and $t$ is the task prompt.

\textbf{Stage 3:} Last, we initialize the conversation again, and instruct the external LLM to expand the single timestamp $t_{best}$ from the last step to a timestamp range. The frame caption at $t_{best}$ is provided, and the LLM is still able to acquire more captions by tool calling to confirm the boundary. Considering the difficulty in dealing with dynamic transitions of events, as explored by some previous work \citep{tfvtg}, we simply apply a hard value of 5 seconds on the output. Formally:
\begin{align*}
(t_{start}', t_{end}') &= stage_3(t_{best}, e, t) \\
(t_{start}, t_{end}) &= (t_{start}' - 5, t_{end}' + 5)
\end{align*}

This three-stage process balances precision and efficiency by minimizing frame access while ensuring robust temporal grounding. For visual clarity of the Isodata clustering, refer to Algorithm~\ref{alg_isodata} for details.

\section{Details on the EMCompress benchmark}
\label{yc2emc_appendix}

We constructed a synthetic question-answering dataset named \textbf{EMCompress}, based on the YouCookII dataset, to support fine-grained temporal and semantic understanding of cooking videos.

\subsection{Source Data Preparation}

The original YouCookII dataset \citep{ZhXuCoAAAI18} contains temporally annotated instructional videos. Each annotation includes a segment $[s_i, e_i]$ representing start and end times (in seconds), along with a natural language description of the cooking step.


To ensure video consistency and avoid duplications, we have verified that each video clip name is unique across the dataset source.

\subsection{Annotation Grouping via Temporal Connectivity}

We define a temporal connectivity criterion to group sequential cooking steps into higher-level event triplets. Given two segments $[s_1, e_1]$ and $[s_2, e_2]$, we define their overlap ratio as:
\[
\text{overlap\_ratio} = \frac{|\min(e_1, e_2) - \max(s_1, s_2)|}{\max(e_1, e_2) - \min(s_1, s_2)}
\]
Two segments are considered \emph{connectable} if:
\[
s_2 > s_1,\quad e_2 > e_1,\quad \text{and}\quad \text{overlap\_ratio} \leq \theta
\]
We set $\theta = 0.1$ in our experiments. Using this rule, we perform a greedy grouping of annotations into connected segments, and extract all valid length-3 subsequences (triplets) from each group.

\subsection{Triplet-Based Question Generation}

Each triplet $T = \{t_1, t_2, t_3\}$ consists of three temporally ordered steps. For each $T$, we generate nine different types of question-answer pairs by instantiating predefined templates. The question types are categorized into three groups:

\paragraph{Temporal Relational Reasoning (TRR)}
\begin{itemize}
    \item \texttt{trr1}: What is the cooking step after of $[description]$?
    \item \texttt{trr2}: What is the cooking step before $[description]$?
    \item \texttt{trr3}: What is the cooking step between $[description]$ and $[description]$?
\end{itemize}

\paragraph{Timepoint Indexed Reasoning (TIR)}
\begin{itemize}
    \item \texttt{tir1}: What is the step between timestamps $s_2$ and $e_2$?
    \item \texttt{tir2}: What is the step between frame indices $f_{s_2} = s_2 \cdot r$ and $f_{e_2} = e_2 \cdot r$?
    \item \texttt{tir3}: What step appears within $f_{d} = (e_2 - s_2) \cdot r$ frames after $s_2$ seconds?
\end{itemize}
Here, $r$ denotes the video frame rate.

\paragraph{Multifaceted Integrative Reasoning (MIR)}
\begin{itemize}
    \item \texttt{mir1}: What is the first step after timestamp $s_2$?
    \item \texttt{mir2}: What is the last step before timestamp $e_2$?
    \item \texttt{mir3}: Within $s_1$ and $e_3$, what is (are) the cooking step(s) apart from $[description]$ and $[description]$?
\end{itemize}

Template instantiation is performed by replacing placeholders with actual sentences and timestamps (framestamps) from the triplet.

\subsection{Data Structuring and Metadata}

Each generated data point is stored with the following fields:
\begin{itemize}
    \item \texttt{vid\_name}, \texttt{vid\_fname}: Video ID and filename.
    \item \texttt{vid\_duration}, \texttt{vid\_frame\_rate}: Metadata from video parsing.
    \item \texttt{type}: One of the nine QA types.
    \item \texttt{question}: Instantiated natural language query.
    \item \texttt{answer}: Corresponding ground-truth step description, serving as the ground-truth label for the VideoQA task.
    \item \texttt{gt\_timestamp}: Temporal segment(s) serving as the ground-truth label for our EMC task on video trimming.
    \item \texttt{gt\_rewritten\_query}: Natural language query serving as the ground-truth label for our EMC task on query re-writing.
    
\end{itemize}

We generated a total of $N = 2754$ QA samples, covering all types evenly.

\subsection{Dataset Splitting}

To support evaluation, we partition the dataset into training, validation, and test splits. For each question type, we allocated:
\[
\text{train: } 1926,\quad \text{val: } 270,\quad \text{test: } 558
\]
This roughly follows 7:1:2.

\begin{figure*}[t]
  \includegraphics[width=\textwidth]{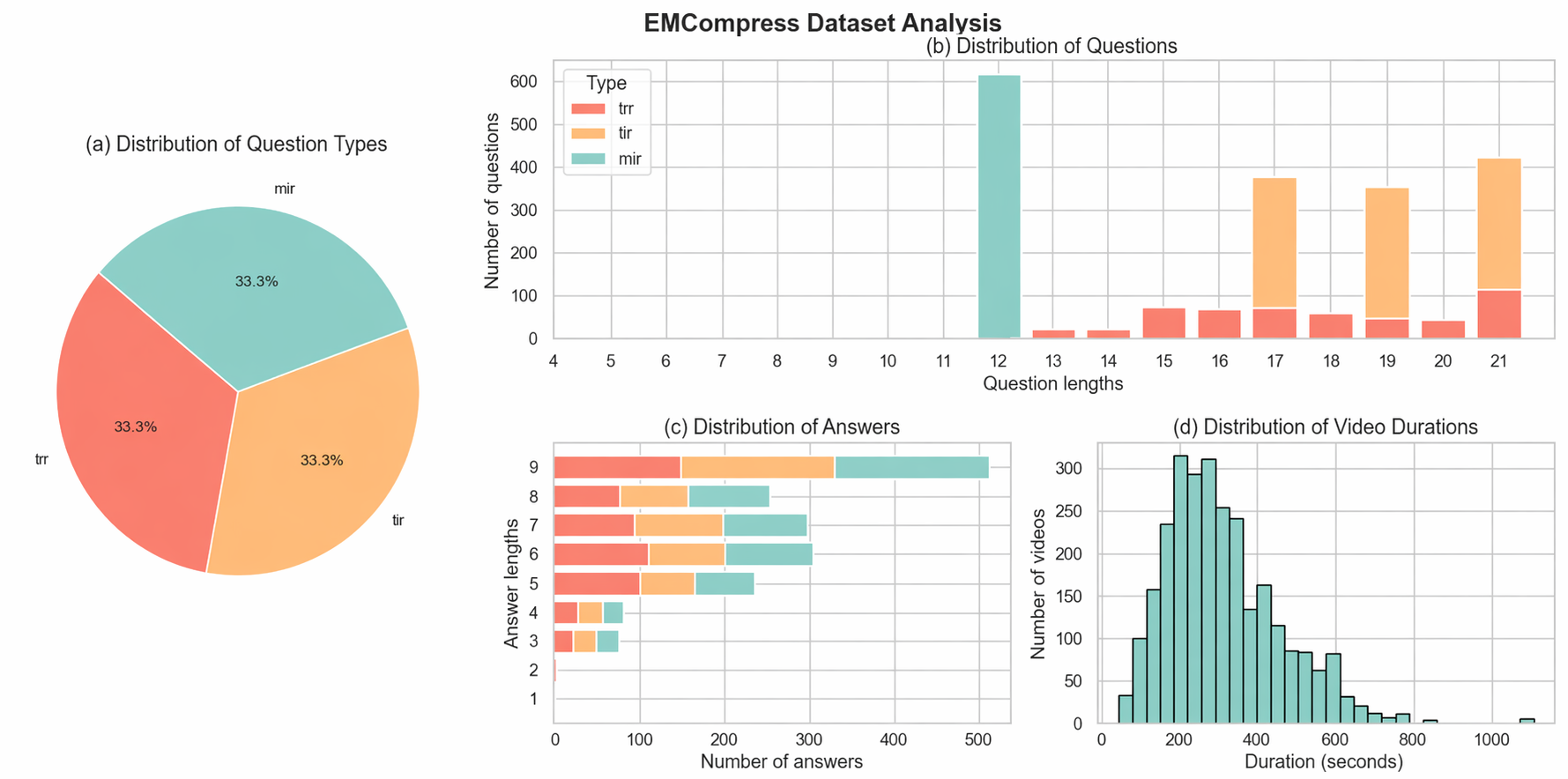}  
  \caption{EMCompress statistics. \vspace{-10pt}}
  \label{fig:yc2emc_statistics}
\end{figure*}

More details about the dataset statistics can be found in Figure \ref{fig:yc2emc_statistics}.

\section{Ablation Study}
\label{ablations}

\begin{table*}[ht]
  \centering

  \begin{subtable}{\textwidth}
    \centering
    \resizebox{\textwidth}{!}{%
      \begin{tabular}{c|cc|cc|cc|cc}
        \toprule
        \textbf{\multirow{2}{*}{Method}}  & \multicolumn{2}{c}{Temporal Relational} & \multicolumn{2}{c}{Timepoint Indexed} & \multicolumn{2}{c}{Multifaceted Integrative} & \multicolumn{2}{c}{\textbf{Average}} \\
         & mIoU & F1 & mIoU & F1 & mIoU & F1 & mIoU & F1 \\
        \midrule
        ReSimplifyIt (Ours) & \underline{0.23} & \textbf{0.37} & \textbf{0.98} & \textbf{0.99} & \textbf{0.47} & \textbf{0.64} & \textbf{0.56} & \textbf{0.67} \\
        ReSimplifyIt-simple (Ours) & \textbf{0.24} & \textbf{0.37} & \textbf{0.98} & \textbf{0.99} & \underline{0.42} & \underline{0.57} & \underline{0.55} & \underline{0.64} \\
        ReSimplifyIt-blind (Ours) & 0.12 & 0.20 & 0.97 & \textbf{0.99} & 0.38 & 0.55 & 0.49 & 0.58 \\
        \bottomrule
      \end{tabular}
    }
    \caption{Results on video output.}
    \label{tab:emc_main_reduced_vid_ablation}
  \end{subtable}

  \begin{subtable}{\textwidth}
    \centering
    \resizebox{\textwidth}{!}{%
      \begin{tabular}{ccccc}
        \toprule
        \textbf{Method} & Temporal Relational & Timepoint Indexed & Multifaceted Integrative & \textbf{Average} \\
        \midrule
        ReSimplifyIt (Ours) & 66.8 & 78.5 & 72.8 & \textbf{72.7} \\
        ReSimplifyIt-simple (Ours) & 66.2 & 81.9 & 68.1 & 72.0 \\
        ReSimplifyIt-blind (Ours) & 65.0 & 80.5 & 70.7 & \underline{72.1} \\
        \bottomrule
      \end{tabular}
    }
    \caption{Results on query rewriting.}
    \label{tab:emc_main_reduced_text_ablation}
  \end{subtable}

  \caption{Ablation studies on our ReSimplifyIt framework. ReSimplifyIt-simple and ReSimplifyIt-blind represent ablations on modular design and feedback from video access, respectively.}
  \label{tab:stage1_eval_ablation}
\end{table*}

\begin{figure*}[t]
  \includegraphics[width=\textwidth]{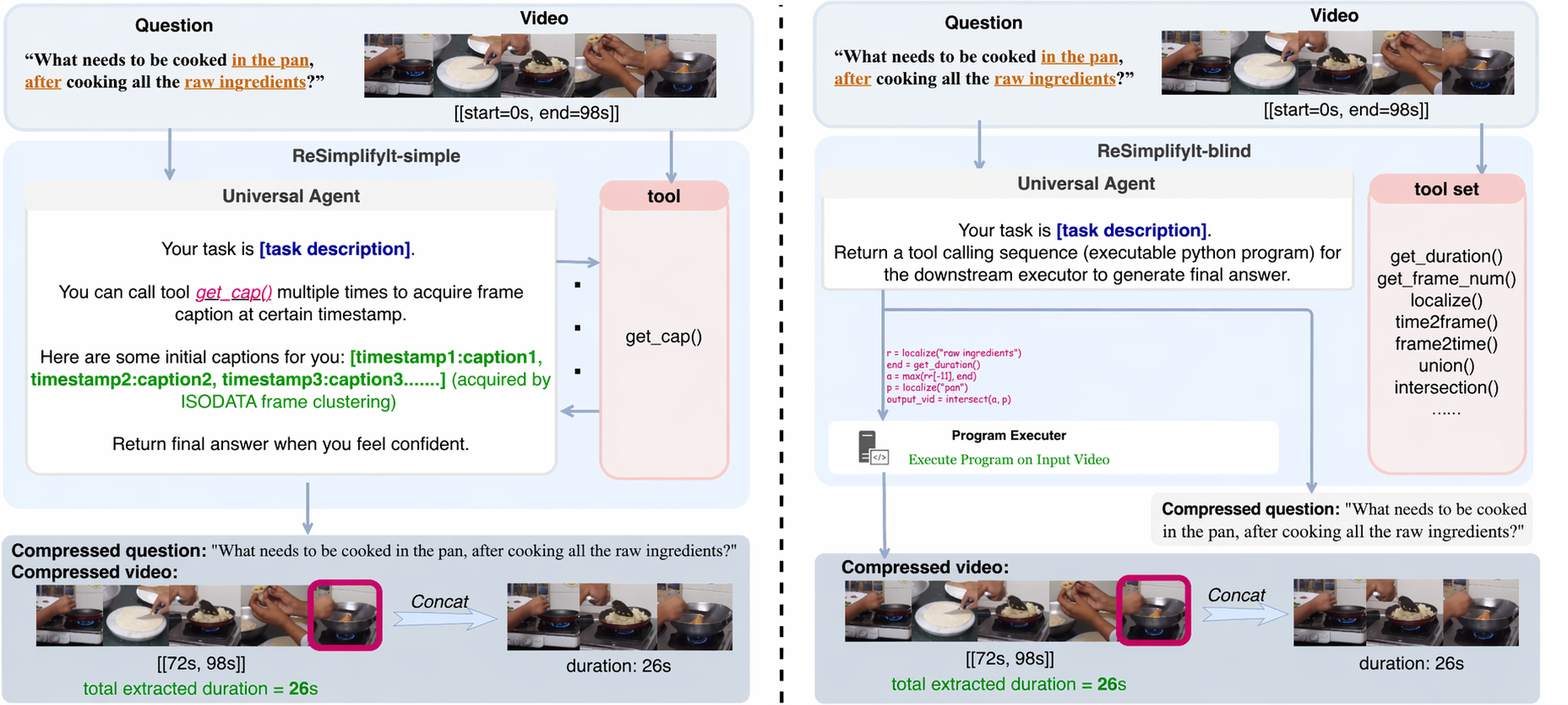}
  \caption{Workflow of \textit{ReSimplifyIt-simple} (left) and \textit{ReSimplifyIt-blind} (right).}
  \label{fig:workflow_figure_ablations}
\end{figure*}

\subsection{Experiment Settings}

We first ablate the modular design—responsible for facilitating structured reasoning—by proposing the \textbf{ReSimplifyIt-simple} variant. In this setting, all underlying interfaces for handling textual and visual inputs, such as ISODATA-clustering (Algorithm~\ref{alg_isodata}) and frame-caption querying, are preserved. However, the entire reasoning pipeline is collapsed into a single, universal agent. Specifically, we initialize an external agentic LLM with task descriptions and operational instructions, and expose the aforementioned interfaces either through conversational context or tool invocation. This unified agent is then responsible for all reasoning procedures—functionally covering the roles of the Launcher, Validator, and Viewer modules, as well as memory tracking—in the original ReSimplifyIt framework, and ultimately produces the final output.

Then, we further ablate the video access completely, i.e., no access to the video content is provided throughout the \textit{entire} reasoning process of the external agent. As the multi-turn interactions are essentially references to video information for refining text input, disabling video access also renders multi-turn interactions unnecessary. To reflect this, we introduce the \textbf{ReSimplifyIt-blind} variant, in which a tool-calling LLM generates a rewritten query and a fixed sequence of tool invocations within a single-turn conversation. This sequence is subsequently executed by a separate executor module to produce the video output.

Refer to Figure \ref{fig:workflow_figure_ablations} for an illustration of these frameworks.

\subsection{Evaluation Results}

Evaluation results are presented in Table~\ref{tab:stage1_eval_ablation}. Although the removal of modular design yields comparable video outputs, the modular architecture still demonstrates advantages—particularly on the \textit{Multifaceted Integrative} subset, which involves more complex multi-hop reasoning, demonstrating the effectiveness of modular design and structured reasoning in more complex and intricate reasoning scenarios. In comparison, the ablation of video access brings notable performance drop, underscoring the critical role of cooperative, feedback-driven multi-turn interaction. This aligns with the interdependency of the vision and text modality, which lies at the core of our EMC task formulation, reflecting the essence of the task initiative.

\paragraph{A performance--cost spectrum of EMC instantiations.}
Taken together, ReSimplifyIt, ReSimplifyIt-simple, and ReSimplifyIt-blind expose an explicit performance--cost spectrum of EMC rather than a single operating point. The full ReSimplifyIt prioritizes robustness and exploratory reasoning through iterative multi-agent interaction, at the cost of more tool calls, captions, and LLM turns. ReSimplifyIt-simple flattens the interaction structure into a single unified agent to substantially reduce overhead, at a small accuracy cost. ReSimplifyIt-blind further removes all video access during compression, representing the lowest-cost extreme in which the trimming plan is produced entirely from the language query. This spectrum demonstrates that EMC can be configured either for efficiency-oriented deployment (Simple/Blind) or for maximal reasoning capability (Full) depending on application requirements, and that the observed gains arise from the compression mechanism itself rather than from any fixed amount of inference compute. Section~\ref{sec:efficiency} quantifies the cost characteristics of each variant across benchmarks.

\subsection{Full list of tools for ReSimplifyIt-blind framework}
\begin{lstlisting}
1. get_duration():
   Return the duration of the video as a floating point value.

2. get_resolution():
   Return the resolution of the video, as a tuple.

3. get_total_frame_num():
   Return total number of the frames of the video, as an integer.

4. grounding_select(obj_name, concerned_indices_input):
   Return, in the form of a list of integers, the indices of all frames
   containing the object given by obj_name, after taking the intersection
   of indices provided by concerned_indices_input. If None is passed,
   selects all frames.

5. indices_list_intersect(list1, list2):
   Return the intersection of two lists of indices.

6. indices_list_union(list1, list2):
   Return the union of two lists of indices.

7. indices_concat_and_fill(list1, list2):
   Return the sorted union of list1 and list2, then fill in missing
   values to make the sequence continuous.

8. indices_concat(list1, list2):
   Return the concatenation of the two lists.

9. timestamp_to_single_index(timestamp):
   Return the frame index corresponding to the given timestamp (in seconds).

10. single_timestamp_to_index_range(timestamp):
    Return indices of 60 consecutive frames centered at the timestamp.

11. range_timestamp_to_index_range(start, end):
    Return all frame indices between the start and end timestamps.
\end{lstlisting}

\section{Implementation details of ReSimplifyIt framework}
\label{appendix:imp}
The Launcher module is based on single-turn conversation with GPT-4o. The Validator and Viewer module, including the scanner and localizer, are primarily based on multi-turn conversation with GPT-4o. For ISODATA frame clustering, we use CLIP \citep{clip} to obtain the visual features of the I-frames. We adopt the off-the-shelf LLaVA-1.5 \citep{Liu_2024_CVPR} for frame captioning. 

\section{Prompts}
\label{prompts}

\subsection{Prompt used for evaluation of query rewriting of EMC}

\begin{lstlisting}
You are a helpful assistant to evaluate the quality of an output of a special type of sentence compression, which sentence is in the form of a question.

You will be given an output of such compression process and the ground truth answer, and the original input question sentence before the compression. Please evaluate the output based on the following three criterias:

    1. Relevance: to minimize unwanted information
    - in this criteria, a candidate output gets full mark if it doesn't contain any information (phrases, concepts, etc.) out of the scope of the original input question.
    - the more information it contains that is not mentioned in the original input question, the more marks are deducted.

    2. Simplicity: to minimize tangential information
    - in this criteria, a candidate output gets full mark if doesn't contain any information (phrases, concepts, etc.) that is included in the original input question but not included in the ground truth compressed question. In other words, whether the question sentence is fully compressed.
    - the more information it contains that is included in the original input question but not included in the ground truth question, the more marks will be deducted.

    3. Completeness: to minimize over-compression
    - in this criteria, a candidate output gets full mark if it contains all information included in the ground truth compressed question. 
    - the more information contained in the ground truth output question is found missing in the output compressed question, the more marks will be deducted. 

    Here is the original input question: [original_question_flag], and
    here is the ground truth compressed question: [ground_truth_compressed_question_flag], and
    here is the output compressed question: [output_compressed_question_flag].

    Please rate the output compressed question on a scale from 0 to 100, with 0 being the worst and 100 being the best (full mark). 
    Now, rate the quality of the output compressed question based on all the information above. 
    Return your answer in this json format: {"score": [your score, from 0 to 100]}.
\end{lstlisting}

\subsection{Prompts used for ReSimplifyIt framework}
Prompt for the Launcher module:
\begin{lstlisting}
Given a video question answering case, i.e. a video, a question, and an answer, sometimes it is possible to cut the video by only keeping a sub-clip or several sub-clips (concatenate if so) to be the video output and simultaneously modify the text question to be the paired text output while keeping the answer consistent and unchanged, so that the answer is still completely compatible to the modified question and the obtained video clip. 
We define this action as "compression". By doing compression, the video becomes shorter, and therefore introduces lower cost to the downstream video question answering model. When modifying the question, note that the downstream video question answering model will only be seeing the sub-clip and think that the shown sub-clip is the whole video, so make sure the modified question is perfectly paired and aligned with and adapted to the modification. 

[In context examples flag 1]

Sometimes, such compression can be repeatedly applied sequentially by several rounds, where the input in each round is the output video clip and the output modified question of its preceeding round. As mentioned, the answer consistency should be always ensured throughout the whole process, and is always completely compatible to the resulting video clip and modified question of every round. If succeeded, each round would make the video shorter and making the situation closer to the optimal case.

[In context examples flag 2]


Your duty is to help complete the compression. Due to some limit, you are only provided with the text question but not the video input.
Your task is to initiate an immediate plan for the compression operation, including a natural language instruction telling which video clip(s) to keep (similar to examples above) and the paired modified text question. If you think the compression may compose more than one round, you only need to perform one round next.
As you cannot see the actual video, your plan might fail, if the downstream video editting agent taking and executing your instruction on video clipping finds it infeasible. Therefore, your plan is refered to as a 'trial'. 
In your current situation, some rounds or trials of the compression process might have already been conducted, and privided as following information:
1, the 'failure history': it is about the history of failed previous trials of the compression of your current round. Here is the failure history for you (an empty indicates that you are making the first trial of this round): [failure_history_flag];
2, the 'success history': it is about the history of the one success trial of all previous round. Here is the failure history for you (an empty indicates that you are at the first round): [success_history_flag].

Now, here is the original question of this round for video question answering: "[quesion_flag]"
Again, please tell your plan which describes what part of the video should be kept. Also, give the modified question under the assumption that the video is processed smoothly according to your plan.

If you feel that there is no room to make such compression (e.g. when the question is being general like "what is this video about") so you feel that you shouldn't make any plan, you should decide to terminate the process.


Hints:
1. Before processing, remember to take a look in the 'failure history' and 'success history' information;
2. If you find that the success history contains a case whose modified_question and the description both significantly overlap with the ones you are about to make, then you should avoid making the same plan again. In this case, you should switch to a clear reasonable sensible alternative plan, or make a decision to terminate the modification process if you can't confidently find one;
3. If you find that the failure history contains a case whose modified_question and the description both significantly overlap with the ones you are about to make, or that any of the "reason" in the failure history records is going to make your attempt fail, then you should avoid making the same plan again. In this case, you should switch to a clear reasonable sensible alternative plan, or make a decision to terminate the modification process if you can't confidently find one.

Return your plan in this json format (keep in mind here, that your response should be in json format):
{"decision": [your decision, either "process" or "terminate"], "modified_question": [the modified question, or "N/A" if your previous decision is "terminate"], "description": [Description of what part of the video should be kept as wanted sub-clip. The description will be passed to downstream processer to validate. Return "N/A" if your decision is "terminate".]}

"""
\end{lstlisting}

Prompt for the Validator module:

\begin{lstlisting}

You will be given a natural language instruction telling you to trim a video, which instruction itself might be infeasible. The reason it might be infeasible is because the agent who gave the instruction had no access to the actual video content, so it might be infeasible if take the actual video contenet into consideration.

Therefore, I need you to be a helpful assistant to confirm if the trimming plan is feasible. 
Specifically, your job is to act as a validator to validate whether it is feasible (whether the video content really supports the plan). 
If it is feasible, you will need to implement the plan and return the resulting sub-clip of the plan in the form of a two-layer list. [In context example flag 1]
If it is not feasible, you will need to tell the reason.

Here are some examples and explanations for infeasible trimming instructions:

[start of examples]

[In context examples flag 2]

[end of examples]

You can invoke a viewer multiple times to acquire the video content (partial content each time), which viewer is a downstream module prepared to assist you.

Note that the viewer is capable to deal with two types of questions:

1. snippet rough scanning, e.g. "what is the video about from xxx second to xxx second"?
2. localizing, e.g. "which segment contains xxx (event/object)?"

Therefore, if [your decision] is "view", [your message] should follow one of the above two example templates. 

Here is the trimming instruction you need to validate: [plan_flag]. 
The video length is [video_length_flag] seconds, and its frame rate is [frame_rate_flag] fps.

Now, in each following turn of this conversation, you need to give your response in this json format: {"decision": [your decision], "message": [your message]}. This works as follows:
[your decision]: either "succeeded", "failed", or "view". "succeeded" means that the plan is successfully implemented, "failed" means that it is not, and by "view" you invoke the viewer to provide partial video content for you. If you choose "view", the viewer will take your message and return as you requested, and the conversation will continue. If you choose "succeeded" or "failed", it will be your final decision and the conversation will end.
[your message]: if you choose "succeeded" as your decision, then it should be the two-layer list as mentioned before, as the video edit result of the plan. If you choose "failed", this should be a brief reason on the failure (e.g. requested timestamp exceeds video length, video doesn't have the object/event needed, etc.). If you choose "view", this should be the question to ask the viewer about the video content.

Hint: it is not always necessary to invoke the viewer. [In context example flags 3]

For your ease of decision, here are some initial frame captions and their timestamps for you (in the form of key value pairs, where the value is the frame caption and the key is its corresponding timestamp, in the unit of second): [initial_captions_flag].
[sys_usr_split]

Now let's start!

\end{lstlisting}

Prompt for Viewer module:

\begin{lstlisting}

You are a helpful and smart assistant that can respond to an upstream request about a video by invoking tools. The length of this video is [video_length_flag].

Here is the content of the request: [validator_request_flag].

Here are the tools you can access (you might access them multiple times if you want):

1. scan(start, end): Return the overall caption of the video snippet (clip) between start and end timestamp, which are the parameters with the unit of second.
2. localize(query): Return the video location, in the form of a timestamp range given by the start and end timestamp, which contains the visual content of the query (which query might be an object, event, etc.)
3. get_image_cap(timestamp): parameter "timestamp" is an integer in the unit of second. Return the caption of the video frame at the given timestamp (regard the frame as an image).

Now, in each turn of the following conversation, your response should be in the following json format: {"decision": [your decision], "message": [your message]}. This works as follows:
[your decision]: either "tool" (if you wish to call tool in this round) or "respond" (if you feel that you are able to respond to the upstream module's request by comprehending your current knowledge acquired about the video.).
[your message]: if your decision is "tool", then this should only contain tool calling following given format, e.g. 'get_image_cap(10)', 'scan(21, 35)', 'localize("kicking the ball")'. If your decision is "respond", then this should be your response to the upstream request.

[sys_usr_split]

Now let's start!

\end{lstlisting}

\subsection{Prompt used for ReSimplifyIt-simple framework}

\begin{lstlisting}

You are a assistant for the video question answering process, in which a candidate is presented with a video and a question for them to answer.\
Your objective is to help the candidate so that they will be able to give the answer with watching the shortest posible sub-clip(s) of the video. \
Your task is to cut the video to acquire this sub-clip(s) and also to modify the question, so that the candidate directly answering your modified question with presented only this sub-clip(s) of the video would be equivalent to answering the original question with presented the original whole uncut video. \

[In context examples flag 1]

You will need to cut the video in the form of providing me the timestamps, which is a list of [start, end] unit clips in the unit of second. \
A tool (python function) will be helping you to get the frame caption of at a certain timestamp (in the unit of second). Whenever you need to call this tool, send a message in this json format: {"decision": "tool", "parameter": [timestamp you need]}. [In context example flag 2] \
Whenever you think you are confident enough to provide the timestamp, return {"decision": "end", "timestamps": [your result timestamps], "revised_question": [your revised question]}. [In context example flag 3].

Before we formally begin, here is a set of original captions with their timestamps provided for you to have an overall rough understanding of the video: [initial_captions_flag].
Also, the frame rate of this video is [frame_rate_flag] frames per second, and the total duration is [duration_flag].

[sys_usr_split]Now let's begin! and the original question is "[original_question_flag]".

\end{lstlisting}

\subsection{Prompt used for ReSimplifyIt-blind framework}

\begin{lstlisting}
You are a assistant for the video question answering process, in which a candidate is presented with a video and a question for them to answer.\
Your objective is to help the candidate so thattheywill be able to give the answer with watching the shortest posible sub-clip(s) of the video. \
Your task is to cut the video to acquire this sub-clip(s) and also to modify the question, so that the candidate directly answering your modified question with presented only this sub-clip(s) of the video would be equivalent to answering the original question with presented the original whole uncut video. \

[In context examples flag 1]

You will be provided with a list of tools to process the video, and the original question to be answered by the candidate based on which to select the frames. Here is the list of tools you have access to, with the description (content in the brackets are the arguments needed):
[1]: get_duration(): return the duration of the video as a floating point value.
[2]: get_resolution(): return the resolution of the video, as a tuple.
[3]: get_total_frame_num(): return total number of the frames of the video, as an integer.
[4]: grounding_select(obj_name, concerned_indices_input): return, in the form of a list of integers, the indices of all frames containing the object given by obj_name, after taking the intersection of indices provided by the argument 'concerned_indices_input'. 'concerned_indices_input' is also a list of indices, and will be set to indices of all frames in the video if 'None' is passed.
[5]: indices_list_intersect(frame_indices_list_1, frame_indices_list_2): return, in the form of a list of integers, the intersection of the two arguments as list. Both argument 'frame_indices_list_1' and 'frame_indices_list_2' are a list of indices.
[6]: indices_list_union(frame_indices_list_1, frame_indices_list_2): return, in the form of a list of integers, the union of the two arguments as list. Both argument 'frame_indices_list_1' and 'frame_indices_list_2' are a list of indices.
[7]: indices_concat_and_fill(frame_indices_list_1, frame_indices_list_2):  first take the sorted union of the two lists given by the arguments, and then fill in all the missing values so that every two adjacent element only differ by 1. Both argument 'frame_indices_list_1' and 'frame_indices_list_2' are a list of indices.
[8]: indices_concat(frame_indices_list_1, frame_indices_list_2): return, in the form of a list, the concatenation of the two lists provided by the arguments.
[9]: timestamp_to_single_index(timestamp): return a list with a single integer, which integer is the index of the frame at the given timestamp. The argument timestamp is a floating point value, whose unit is second.
[10]: single_timestamp_to_index_range(timestamp): return, in the form of a list, the indices of 60 consecutive frames, the midpoint of which is at the given timestamp. The argument timestamp is a floating point value, whose unit is second.
[11]: range_timestamp_to_index_range(start, end): return, in the form of a list, the indices of all frames which are between the two timestamps which are provided by the arguments. The argument start and end are both floating point values, whose unit are both second.

Above are all the tools to have access to. Please note that selecting frames out of all the frames of the original video is being cut and clipped, therefore you will also need to modify the aforementioned prompt, to make it align well with the reduced video frames.
[sys_usr_split]
Now the original question is: [question_flag]. Having access to the information of all the tools mentioned above, provide me the python code which could achieve the selection of frames. You may define variables to store intermediate result, and determine the value of some arguments when necessary, but you should not require the downstream task operator to replace any of your assumption on arguments, as no more information but the original video is provided to the downstream task. Please use the variable name 'final_frames' to store your final list of frame index. Please only provide the code and revised question in this format: {"Code:[your whole paragraph of code] Revised question:[your revised prompt]"}, where [your whole paragraph of code] should be an empty string if you think no tools need to be called and the whole original video should be passed to the downstream task.
\end{lstlisting}

\section{More on Related Work}
\label{appendix: related_work}

\paragraph{Video-LLMs for VideoQA}

Video-LLMs have spurred a wave of models aimed at enhancing video understanding by leveraging the language capabilities of large language models (LLMs)~\citep{lin2023video, Ma_2024_CVPR, 10658165, Li2023LLaMAVIDAI, liu2024llavanext, Xu2024PLLaVAP, InternVL3, qwen2.5vl, qwen3vl, llavaonevision}. Some approaches~\citep{chen2023video-LLMmodelingvideosequence, 10658165, 2023videochat} utilize dedicated video encoders such as video transformers~\citep{gberta_2021_ICML} or convolution networks. However, these designs often demand large-scale annotated video-text data and significant computational resources. To address this, alternative methods adapt pre-trained image-domain MLLMs to video inputs~\citep{10658172, maaz-etal-2024-video}, offering improved practicality. Nonetheless, the sparsity and query-invariant nature of video encoding in existing models limits their ability to capture fine-grained spatial-temporal details effectively, especially under token budget constraints. This work addresses such inefficiencies by introducing a query-adaptive processing mechanism inspired by the principle of information compression, aiming to reconcile the trade-off between token efficiency and representational fidelity.

\paragraph{LLM-assisted Agentic Reasoning for VideoQA}
Another line of research for video question answering lies in building pure-text LLM assisted frameworks or multi-agent systems for VideoQA \citep{wang2024videotreeadaptivetreebasedvideo, shang-etal-2024-traveler, VideoAgent}. Compared to Video-LLMs, which represents end-to-end single-pass approaches, these methods adopt traditional or LLM-based methods to proactively sample relevant video frames. VideoAgent \citep{VideoAgent} and TravLER \citep{shang-etal-2024-traveler} utilized LLM's planning ability to conduct iterative keyframe searching, while VideoTree \citep{wang2024videotreeadaptivetreebasedvideo} presented query-adaptive hierarchical tree-based keyframe selection. 

\paragraph{Temporal Sentence Grounding for Videos}

Temporal sentence grounding (TSG) aims to localize the video segment best matching a language query. Early sliding-window methods (e.g., TALL~\citep{8237825}, MCN~\citep{8237880}) were costly, while later proposal-based~\citep{xu2019multilevel,10.1609/aaai.v33i01.33018199} and proposal-free methods~\citep{10.1609/aaai.v33i01.33019159,zhang-etal-2020-span} improved efficiency via query-guided proposals or direct boundary prediction. All these methods take a video and a query as input and predict a matching temporal span. In contrast, our proposed EMC generalizes beyond TSG by supporting inter-frame reasoning and joint adaptation over both video and query, rather than retrieving only frames directly mentioned in the query.

\paragraph{Grounded videoQA}
Another seemingly similar line of research lies in integrating grounding techniques into VideoQA pipelines \citep{nextgqa, cgbench}. Grounded VideoQA seeks to enhance model faithfulness by explicitly linking a model's textual output to ``visual cues'' or ``evidence'' within the video. This approach is effective at mitigating hallucinations for descriptive queries by enforcing \textit{perceptual} alignment. While effective for descriptive tasks, the approach suffers from a fundamental conceptual incongruity with the nature of complex video reasoning. The semantics of a query and its corresponding video segment are often holistically entangled; for example, a high-level question about intent, causality, procedure, or even a simple temporal relational reasoning does not map to an atomic piece of visual evidence but is inferred from a continuous temporal context. The attempt to impose a discrete justification framework onto this intertwined semantic space leads to an inherently brittle and ill-defined notion of a ``visual cue''. 

EMC naturally sidesteps this ambiguity by fundamentally reframing the objective, turning from \textbf{\textit{``what to answer''}} to \textbf{\textit{``what to ask''}}. More fundamentally, EMC introduces a more principled paradigm that respects this intrinsic entanglement. Rather than seeking to atomize evidence for a given answer, EMC reformulates the task itself through a priori contextual simplification. It isolates the minimal, self-contained $(video, query)$ sub-problem through an \textbf{endomorphic transformation} that keeps with the original task space unchanged and preserves the necessary holistic context for reasoning. This approach is not merely a circumvention of the definitional challenge; it establishes a more fundamentally sound and cognitively aligned task. By first reducing the problem space to a manageable and semantically coherent unit—mirroring the human strategy of simplifying a problem before attempting to solve it—EMC offers a more robust foundation for complex video-language understanding.

\paragraph{Information Bottleneck}
The Information Bottleneck principle~\citep{tishby1999information, tishby2015deep} formalizes representation learning as a sufficiency--compression trade-off, extended by the variational IB~\citep{alemi2017deep} to deep models. EMC shares the same sufficient-statistic substrate but operates in a different regime along two axes: it compresses over the original multimodal input space rather than a learned latent code, and it realizes sufficiency as a pointwise, model-agnostic condition on task behavior (C2) rather than as a distribution-level regularizer, yielding the interpretable discrete artifact $(v, q)$.

\section{Cost Driver Details}
\label{appendix:efficiency_full}

This section provides per-dataset cost-driver statistics for both EMC instantiations, complementing the headline CostRatio analysis of Section~\ref{sec:efficiency}. Table~\ref{tab:cost_drivers_simple_app} reports the high-level cost drivers (caption count, tool calls, LLM turns, output tokens) and compression effectiveness (DurAll, DurScrn, Compress\%) for ReSimplifyIt-simple. Table~\ref{tab:cost_drivers_full_app} reports the same high-level statistics for the full ReSimplifyIt. Table~\ref{tab:cost_full_breakdown} further decomposes the captions of the full ReSimplifyIt into passive (pre-loaded) versus active (tool-fetched) sources; column definitions follow the list below.

\begin{itemize}[leftmargin=*,noitemsep,topsep=0pt]
    \item \textbf{Pasv}: passive captions per sample ($=\text{VaIn}+\text{LoIn}$), embedded in prompts without explicit tool calls.
    \item \textbf{VaIn}: Validator initial captions, uniformly sampled frames embedded in the Validator's prompt ($\sim$10 per trial).
    \item \textbf{LoIn}: Localize initial captions; each \texttt{localize()} call samples $k{=}10$ uniform frames for its sub-agent prompt.
    \item \textbf{Actv}: active captions per sample ($=\text{\#TotalCap}-\text{Pasv}$), fetched during tool execution via \texttt{scan}, \texttt{localize}, and \texttt{get\_image\_cap}.
    \item \textbf{V$\rightarrow$Vw}: Viewer sessions spawned per sample (number of times the Validator invokes the Viewer as a tool).
    \item \textbf{\#scan}, \textbf{\#localize}, \textbf{\#get\_cap}: the breakdown of total Viewer tool calls (VwTl) into its three constituent tool types.
\end{itemize}

\begin{table}[ht]
  \centering
  \small
  \setlength{\tabcolsep}{4pt}
  \resizebox{\columnwidth}{!}{%
  \begin{tabular}{l|cccc|ccc}
    \toprule
    \textbf{Dataset} & \textbf{\#TotalCap} & \textbf{\#Tool} & \textbf{\#LLM} & \textbf{\#OutTok} & \textbf{DurAll} & \textbf{DurScrn} & \textbf{Compress\%} \\
    \midrule
    ActivityNet-QA  & 21.4 & 1.9 & 3.2 & 341  & 9.8\%  & 9.0\%  & 99.2\% \\
    EMCompress   & 21.9 & 1.9 & 3.3 & 366  & 6.4\%  & 6.4\%  & 100.0\% \\
    NExT-OE         & 23.4 & 3.9 & 5.0 & 433  & 20.2\% & 16.7\% & 95.7\% \\
    EgoSchema       & 22.1 & 2.1 & 3.5 & 446  & 26.0\% & 26.0\% & 100.0\% \\
    LVBench         & 27.0 & 7.0 & 9.6 & 1482 & 2.2\%  & 1.9\%  & 99.7\% \\
    MLVU            & 24.9 & 4.9 & 6.9 & 1122 & 12.2\% & 7.0\%  & 94.4\% \\
    Video-MME       & 23.9 & 3.9 & 5.7 & 806  & 8.3\%  & 8.0\%  & 99.7\% \\
    \bottomrule
  \end{tabular}%
  }
  \caption{Per-dataset cost drivers and compression effectiveness for ReSimplifyIt-simple.}
  \label{tab:cost_drivers_simple_app}
\end{table}

\begin{table}[ht]
  \centering
  \small
  \setlength{\tabcolsep}{4pt}
  \resizebox{\columnwidth}{!}{%
  \begin{tabular}{l|ccc|ccc}
    \toprule
    \textbf{Dataset} & \textbf{\#TotalCap} & \textbf{V$\rightarrow$Vw} & \textbf{VwTl} & \textbf{DurAll} & \textbf{DurScrn} & \textbf{Compress\%} \\
    \midrule
    ActivityNet-QA  & 67.5 & 2.7 & 6.8  & 51.7\% & 16.2\% & 57.6\% \\
    EMCompress   & 73.9 & 2.4 & 5.8  & 6.8\%  & 6.8\%  & 97.9\% \\
    NExT-OE         & 18.3 & 0.4 & 1.3  & 86.9\% & 32.8\% & 19.5\% \\
    EgoSchema       & 50.6 & 2.3 & 4.5  & 75.0\% & 25.6\% & 33.6\% \\
    LVBench         & 83.8 & 2.6 & 11.3 & 57.5\% & 4.8\%  & 44.6\% \\
    MLVU            & 64.5 & 1.7 & 5.2  & 81.4\% & 7.2\%  & 20.1\% \\
    Video-MME       & 76.2 & 2.4 & 6.7  & 64.0\% & 10.4\% & 40.1\% \\
    \bottomrule
  \end{tabular}%
  }
  \caption{Per-dataset cost drivers and compression effectiveness for the full ReSimplifyIt. VwTl denotes total Viewer tool calls per sample.}
  \label{tab:cost_drivers_full_app}
\end{table}

\begin{table}[ht]
  \centering
  \small
  \setlength{\tabcolsep}{4pt}
  \resizebox{\columnwidth}{!}{%
  \begin{tabular}{l|cccc|ccc}
    \toprule
    \textbf{Dataset} & \textbf{Pasv} & \textbf{VaIn} & \textbf{LoIn} & \textbf{Actv} & \textbf{\#scan} & \textbf{\#loc} & \textbf{\#get\_cap} \\
    \midrule
    ActivityNet-QA  & 34.0 & 17.4 & 16.6 & 33.5 & 2.6 & 1.7 & 2.4 \\
    EMCompress   & 38.2 & 22.8 & 15.4 & 35.7 & 3.0 & 1.5 & 1.1 \\
    NExT-OE         & 5.3  & 2.7  & 2.6  & 13.0 & 0.4 & 0.3 & 0.4 \\
    EgoSchema       & 24.5 & 13.6 & 10.8 & 26.1 & 2.5 & 1.1 & 0.8 \\
    LVBench         & 48.9 & 17.5 & 31.3 & 94.9 & 4.8 & 3.1 & 3.0 \\
    MLVU            & 22.1 & 9.8  & 12.3 & 42.4 & 2.1 & 1.2 & 1.6 \\
    Video-MME       & 32.8 & 15.0 & 17.8 & 43.4 & 3.3 & 1.8 & 1.5 \\
    \bottomrule
  \end{tabular}%
  }
  \caption{Full per-source caption breakdown for ReSimplifyIt. \#loc abbreviates \#localize. Passive captions are pre-loaded into agent prompts; active captions are fetched during tool execution.}
  \label{tab:cost_full_breakdown}
\end{table}

\section{EMC Integration into Video-LLM Workflows}
\label{appendix:emc_integration}

We expand the deployment modes summarized in Section~\ref{subsec:emc_integration}. Because $\mathcal{F}_{\text{EMC}}$ is endomorphic, both modes slot into existing pipelines with no architectural change.

\paragraph{EMC for Inference-Time Simplification in Video-LLMs.} At inference, EMC pre-processes $(V, Q)$ into $(v, q)$ (see Section~\ref{sec:method}), reducing the temporal and semantic complexity the downstream model must handle and concentrating it on compact, task-relevant content for fine-grained temporal reasoning and direct visual grounding.

\paragraph{EMC for Improving Training-Time Visual-Language Alignment.}
Video-LLM supervision tuples $(V, Q, a)$ often couple abstract or multi-hop queries with unfocused video. EMC reshapes them into compact triples $(v, q, a)$, where $v$ is a high-utility segment and $q$ is a grounded reformulation of $Q$, offloading off-target supervision to upstream modules so the model specializes in two core competencies: temporal understanding and multimodal alignment.

\section{Additional Discussion}
\label{appendix:discussion}

\subsection{EMC vs.\ Test-Time Reasoning Strategies}
\label{appendix:discussion_orthogonality}

EMC and test-time reasoning strategies such as Chain-of-Thought~\citep{wei2022chain}, Tree-of-Thoughts~\citep{yao2023tree}, and Graph-of-Thoughts~\citep{besta2024graph} address different bottlenecks and are not substitutes for one another. Test-time reasoning strategies primarily modify \textit{how} an LLM explores or restructures its thought process given a fixed input; they do not directly address the sparsity of evidence in long videos, where the dominant failure mode is that a fixed downstream frame budget under-covers the relevant segment to begin with. By contrast, EMC is a front-end endomorphic problem transformation that reshapes the \textit{evidence distribution} itself: by compressing $(V,Q) \to (v,q)$ under answer invariance, it concentrates the same downstream frame budget onto a short relevant segment, yielding the density gains quantified by our $\mathrm{DensAmp}$ and $\mathrm{CostRatio}$ metrics in Section~\ref{sec:efficiency}.

These two directions are therefore orthogonal and complementary: adding reasoning steps does not automatically reproduce EMC's benefit unless it is coupled with explicit evidence selection at the sampling stage, while EMC can be composed with any downstream reasoning strategy. Our framework further exposes a performance--cost spectrum rather than a single operating point (ReSimplifyIt, ReSimplifyIt-simple, and ReSimplifyIt-blind; Appendix~\ref{ablations}), with progressively lower amounts of inference compute. The fact that compression gains persist even in the flattened and blind variants supports that EMC's benefit stems from the evidence-compression mechanism rather than from the sheer amount of LLM compute invested.

\subsection{Minimality as a Desideratum and Boundary Expansion}
\label{appendix:discussion_minimality}

The minimality condition in our formulation (Section~\ref{emc_def}) describes a \textit{target property} of the task formalization---that the output segment $v$ should contain the minimal sufficient visual evidence for answering $q$---rather than a strict guarantee enforced by any particular implementation. In practice, event boundaries in real-world videos are often ambiguous or gradual, and strict minimal cropping is a known failure mode: small boundary errors can easily exclude critical visual evidence and lead to disproportionately large downstream reasoning failures. Our Localizer (Appendix~\ref{appendix_viewer}) therefore applies a conservative temporal expansion of $\pm p$ seconds around the predicted boundary as a safety margin. Condition (C2) is a hard admissibility requirement, not a trade-off dimension; but the Localizer's boundary predictions are stochastic, and overly aggressive crops can drop frames whose removal would in fact violate (C2) on some downstream model. The $\pm p$ margin is therefore a \textit{robustness buffer} for enforcing (C2) under localization uncertainty: smaller $p$ risks genuine (C2) violations (under-inclusion of $A$-relevant evidence), while larger $p$ merely loosens the minimality objectives (O1--O2) by retaining redundant content without affecting admissibility. The chosen $p{=}5$ is the operating point at which (C2) is reliably enforced without excessive redundancy.

\paragraph{Sensitivity to the expansion margin.}
To validate that $p{=}5$ is not arbitrary, we vary $p \in \{1, 3, 5, 7\}$ and report the resulting average mIoU on EMCompress:

\begin{center}
\begin{tabular}{cc}
\toprule
$p$ (seconds) & Average mIoU \\
\midrule
1 & 0.51 \\
3 & 0.53 \\
\textbf{5} & \textbf{0.56} \\
7 & 0.49 \\
\bottomrule
\end{tabular}
\end{center}

We observe a clear trade-off: too small a margin (1--3s) risks missing event boundaries and lowers overlap with the ground truth, while too large a margin (7s) unnecessarily enlarges the segment and reduces localization precision. A moderate margin of $5$ seconds achieves the best balance, and we adopt it as the default in all reported results. An adaptive, confidence-aware or boundary-aware expansion strategy would be an interesting direction for future work.

\subsection{Constraint Re-allocation and the Collapse of the Pareto Frontier}
\label{appendix:pareto_collapse}

A priori, the video-side minimality (O1) and query-side minimality (O2) in Section~\ref{emc_def} admit a Pareto trade-off, so the optimum could span an entire frontier rather than a unique point. \textbf{Note that this concerns the O1--O2 minimization only; admissibility (C1--C2) remains a hard constraint throughout and is not part of the Pareto structure.} In EMC, however, the two minimality objectives are not independent---they are two faces of a single underlying compression act. Each reasoning step removable from $q$ corresponds to a constraint in the original query that can be \textit{structurally pre-satisfied} by trimming $V$: rather than verifying the constraint textually, the transformation excises the non-compliant portion of the video. Consequently, every reduction in $\mathrm{Infer}(q)$ is paid for by an equivalent reduction in the span of $V$ that must be retained to remain $A$-sufficient, and vice versa. The two objectives descend in lockstep along the admissible region, and the Pareto frontier collapses to essentially a single operating point.

We emphasize that this re-allocation is a \textit{mechanism-level account} of why a unique optimum exists; it does not redefine the task. The answer $A$ and the answer space are preserved throughout---no constraint is discarded, each is merely \textit{conserved across modalities}---so EMC remains a joint multimodal transformation, distinct from unimodal query relaxation or evidence-retrieval paradigms~\citep{nextgqa, cgbench}.

This mechanism further justifies the \textbf{video-priority lexicographic} resolution adopted in Section~\ref{emc_def}: video compression is the \textit{driver} of the transformation and query adaptation its downstream consequence---once $v^*$ is fixed, the remaining freedom in $q$ is exhausted by rewriting to the shortest form compatible with $v^*$.

\subsection{On Measurability of the Information-Theoretic View}
\label{appendix:measurability}

The mutual-information terms in $I((v,q); A) \leq I((V,Q); A)$ articulate EMC's membership in the sufficient-statistic family; they are a \textit{theoretical characterization} of the task, not a computable training signal to be estimated numerically. In practice, sufficiency is realized through the VideoQA-natural condition (C2), and minimality is tracked through the computable proxies $\mathrm{Size}(v)$ and $\mathrm{Infer}(q)$. The information-theoretic view thus provides the \textit{language} of the formulation, while the concrete conditions (C1--C2) and objectives (O1--O2) provide its operational content.

\subsection{On the Link between Mutual Information and Answer Invariance}
\label{appendix:mi_answer_link}

The use of mutual information to capture answer preservation rests on the classical sufficient-statistic property: $I((v, q); A) = I((V, Q); A)$ if and only if there exists a decision rule on $(v, q)$ achieving the same Bayes-optimal performance on $A$ as any decision rule on $(V, Q)$~\citep[Ch.~2.9]{cover2006elements}. MI equality is therefore a population-level characterization---the existence of a sufficient-capacity model whose predictions are invariant across the compression. Condition (C2) is the \textit{VideoQA-natural instantiation} of this same sufficiency property: because VideoQA has no distribution-averaged notion of correctness, sufficiency must hold pointwise across all reasonable $\mathcal{M}$, not merely in expectation. The pointwise form implies the MI equality by preserving the full distribution of model outputs, so (C2) is simultaneously a concrete behavioral condition and a realization of the sufficient-statistic regime in the VideoQA setting.

\end{document}